\documentclass{svjour3} 
\smartqed 
\usepackage[utf8]{inputenc}
\usepackage{amssymb}
\usepackage{listings}
\usepackage{mathrsfs}
\usepackage{times}
\usepackage{float}
\usepackage{color}
\usepackage{setspace}
\usepackage{color,soul}

\usepackage{amsmath,amsfonts,amsthm,eucal}
\usepackage{enumerate}
\usepackage[letterpaper,margin=3cm]{geometry}
\usepackage{float}
\usepackage[title]{appendix}
\usepackage{color}

\usepackage[
pdfstartview=XYZ,
bookmarks=true,
colorlinks=true,
linkcolor=blue,
urlcolor=blue,
citecolor=blue,
pdftex,
bookmarks=true,
linktocpage=true,   
hyperindex=true
]{hyperref}

\usepackage{lipsum}
\usepackage{lineno}
\usepackage[FIGBOTCAP,TABTOPCAP,bf,tight]{subfigure}

\usepackage{natbib}
\usepackage[pdftex]{graphicx}
\usepackage{epstopdf}
\usepackage{booktabs} 

\usepackage{tikz,mathpazo}
\usetikzlibrary{shapes.geometric, arrows}

\usepackage{caption}
\usepackage{mathtools}
\usepackage{etoolbox}

\newcommand{\tensor}[1]{\ensuremath{\boldsymbol{#1}}}

\DeclareMathAlphabet{\mathpzc}{OT1}{pzc}{m}{it}

\DeclareMathOperator*{\argmin}{arg\,min}

\theoremstyle{remark}

\renewcommand{\vec}[1]{\ensuremath{\boldsymbol{#1}}}

\usepackage[super]{nth}
\usepackage{array}
\usepackage{multirow}

\newcommand{\norm}[1]{\left\lVert#1\right\rVert}

\newcommand{\tpde}{\text{pde}}
\newcommand{\tdbc}{\text{dbc}}
\newcommand{\tnbc}{\text{nbc}}
\newcommand{\taux}{\text{aux}}
\newcommand{\tnn}{\text{nn}}
\newcommand{\tfem}{\text{fem}}
\newcommand{\tcompy}{\text{compy}}

\usepackage{algorithm}
\usepackage[noend]{algpseudocode}

\title{Training multi-objective/multi-task collocation physics-informed neural network with student/teachers transfer learnings}  

\begin{document}

\titlerunning{Enhanced transfer learning for PINN with labeled data}

\author{Bahador Bahmani      \and 
        WaiChing Sun 
}

\institute{Corresponding author: WaiChing Sun \at
Associate Professor, Department of Civil Engineering and Engineering Mechanics, 
 Columbia University , 
 614 SW Mudd, Mail Code: 4709, 
 New York, NY 10027
  Tel.: 212-854-3143, 
  Fax: 212-854-6267, 
  \email{wsun@columbia.edu}        
}

\date{Received: \today / Accepted: date}

\maketitle

\begin{abstract}
This paper presents a PINN training framework that employs (1) pre-training steps that accelerates and improve the robustness of the training of physics-informed neural network with auxiliary data stored in point clouds, (2) 
a net-to-net knowledge transfer algorithm that improves the weight initialization of the neural network 
and (3) a multi-objective optimization algorithm that may improve the performance of a physical-informed neural network with competing constraints.
 We consider the training and transfer and multi-task learning of physics-informed neural network (PINN) as multi-objective problems where the physics constraints such as the governing equation, boundary conditions, thermodynamic inequality, symmetry, and invariant properties, as well as point cloud used for pre-training can sometimes lead to conflicts and necessitating the seek of the Pareto optimal solution. In these situations, weighted norms commonly used to handle multiple constraints may lead to poor performance, while other multi-objective algorithms may scale poorly with increasing dimensionality. To overcome this technical barrier, we adopt the concept of vectorized objective function and modify a gradient descent approach to handle the issue of conflicting gradients. Numerical experiments are compared the benchmark boundary value problems solved via PINN. The performance of the proposed paradigm is compared against the classical equal-weighted norm approach.  Our numerical experiments indicate that the brittleness and lack of robustness demonstrated in some PINN implementations can be overcome with the proposed strategy. 
\end{abstract}


\section{Introduction}
\label{intro}
 
The seminal works of physical-informed neural network (PINN), 
the procedure that trains neural network to generate solutions of boundary value problems,
has recently gained significant attention presumably due to the breakthrough and popularity of deep learning 
 and the accessibility of the open-source machine learning software such as Tensorflow and PyTorch  \citep{lagaris1997artificial, lagaris1998artificial, mcfall2006artificial, raissi2019physics}. 
A rapidly growing body of literature has shown that the physics-informed neural network 
is capable of solving a large variety of boundary value problems, 
including but not limited to time-dependent PDEs
\citep{meng2020ppinn}, Lagrangian particle-based fluid dynamics \citep{wessels2020neural}, 
phase field fracture \citep{goswami2020transfer}, elasticity \citep{haghighat2021physics}, 
and turbulence flow \citep{duraisamy2019turbulence, li2020fourier}. 
These boundary value problems are often analyzed via more 
conventional PDE solvers (e.g. finite volume, finite element, smoothed particle hydrodynamics)
with decades of development tailoring to guarantee the stability, convergence, and robustness of the 
solutions \citep{oden2003research, hughes2012finite}. 

The Physics informed neural network is distinctive from other machine learning paradigms commonly used for mechanics and physics problems in the way data is demanded and used. 
Unlike the supervised learning commonly used for materials laws where
 artificial intelligence (e.g. neural network, support vector machine)  are trained with labels (e.g. stress-strain, porosity-permeability-formation-factor, equation of state) to make forecasts \citep{cai2021equivariant, wang2018multiscale, tartakovsky2018learning, fujimoto2018methodology, logarzo2021smart, nassar2019modeling}, the search of the solution itself in PINN does not demand any data except those required to constitute the loss function. While unsupervised learning that determines lower-dimensional features and structures to represent high-dimensional data could be helpful to improve the performance of the training, it is not a necessary ingredient for PINN. 
A Google search on Physics informed neural network may yield more than 90,000 results as of July 2021, which indicates the massive
efforts invested in this rapidly growing paradigm \citep{raissi2019physics, wang2021eigenvector, qian2020lift}. 

Nevertheless, the training of the physics-informed neural network is by no means trivial. First, unlike a more established framework such as finite element method that has 
clear strategies (e.g. mesh refinement) and established mathematical analysis to guarantee convergence and stability for the pre-determined linear finite-dimensional spaces for the solution and weighting functions \citep{szabo1991finite, ainsworth1997posteriori}, the multi-layer perceptron is built by alternating linear transformation and nonlinearities applied element-wise to the output of the linear transformation \citep{montavon2011kernel}. Furthermore, there also exhibit multiple ways to express the physical constraints or governing equations for 
both forward and inverse problems including the collocation-based loss function \citep{dissanayake1994neural,lagaris1998artificial,sirignano2018dgm,raissi2018deep,raissi2019physics, xu2019neural}, which evaluate the solution at selected collocation points, and the energy-based (Ritz-Galerkin) method
that requires numerical integrations but also reduces the order of the derivatives in the governing equations, 
 \citep{weinan2018deep,samaniego2020energy}, and the related variational approach \citet{kharazmi2021hp} that parametrizes trial and test spaces by neural network and polynomials, respectively. This vast number of choices is further complicated by the large number of 
tunable hyperparameters, such as the configurations of the neural network \citep{fuchs2021dnn2, heider2021offline, psaros2021meta}, the types of activation functions \citep{psaros2021meta} and the neuron weight initialization \citep{glorot2010understanding,he2015delving,goodfellow2016deep,cyr2020robust}, and different techniques to impose boundary conditions \citep{sukumar2021exact} , while providing significant flexibility, also could make 
it confusing for researchers unfamiliar with neural network to determine the optimal way to train the PINN properly and efficiently. 
Finally, the recent work on meta-learning and analysis on the enforcement of boundary conditions also reveals that the multiple physical constraints employed to measure and minimize the error of the approximated solution may lead to conflicting situations where 
actions that reduce one set of constraints may also increase the error of another set of constraints and lead to a comprised local minimizer that is not desirable \citep{psaros2021meta, yu2020gradient, rohrhofer2021pareto}.  

This paper focuses on the training aspect of the physics-informed neural network. 
Our goal is to explore ways to make the training of PINN less precarious, reduce the number of trial-and-error required to properly tune the hyperparameters, and at least empirically improve the robustness of the training process. We limited the scope of this study to the collocation physics informed neural network, although the techniques presented in this paper may be applicable to the other variants. 
Our new contribution here is to introduce a synthesis that includes (1) a pre-training step where solutions from cheap solvers or spatial data from experiments may provide labeled data to facilitate the pre-training that precede the training of PINN constraints, (2) the use of Net2net transfer learning that may reduce the dimensionality of the hyperparameter training and (3) the gradient surgery method that enables one to handle the conflicting gradients coincide with high positive curvature and significant difference in magnitude of the gradient components that might otherwise jeopardize the training. Our numerical experiments show that this combination of strategies together may significantly simplify the training of PINN in a selected boundary value problems that are known to be difficult to solve directly via PINN in the literature.

\subsection{Literature review on multi-task learning relevant for PINN}
The optimization problem concerning multi-objective multi-task learning with deep neural network is an ongoing research topic where numerous methods have been proposed for many applications both supervised and reinforcement learning \citep{ruder2017overview, yang2019generalized}. One common strategy is to define a scalarized total loss function that is a weighted product norm of task/objective-specified loss functions. For instance,  GradNorm (cf. \citet{chen2018gradnorm}),  Gradient Surgery (cf. \citet{yu2020gradient}) and the recent work on meta-learning for loss function (cf. \citet{psaros2021meta}) introduce different remedies,  to 
either adjusting the task weight adaptively (and hence varying the learning speed of each sub-task)  or modify the search direction during the training to improve the balance among different tasks 
and find the Pareto-optimal solution. 
Meanwhile, other possible alternatives include treating these task weights as extra hyperparameters (that are tuned via an exhaustive search,) and empirically designing specific neural network architecture (e.g. the cross-stitch network \citep{misra2016cross}) to accommodate the multi-task optimization.

\citet{kendall2018multi} considers tasks' weights as additional trainable parameters to control homoscedastic uncertainty associated with each task. \citet{chen2018gradnorm} utilizes the gradient norm of each task's loss function as an indicator to appropriately balance the total gradient vector among different tasks such that they can be improved with the same rate during the gradient descent iterations. A Frank-Wolfe-based optimizer is proposed by \citet{sener2018multi} that aims to address conflicting objectives issues by finding the optimal update direction among task-specific gradient vectors. A simple projection technique is introduced by \citet{yu2020gradient} that modifies only gradient vectors of those objectives that are in conflict with each others. 
One concern with the schemes that are dealing with task-specific gradient calculations is the extra computational cost due to the gradient calculations. Another class of methods tries to utilize statistical properties of task-specific loss functions as a mean to estimate optimal weights  without direct access to their gradient vectors \citep{kiperwasser2018scheduled,jean2019adaptive,leang2020dynamic,groenendijk2021multi}.

In our proposed model, we have broken down a machine learning task to generate PDE solution into a multi-task multi-objective optimization problem. The introduction of the pre-training step that employs supervised learning of labeled data is considered as auxiliary tasks where coarse knowledge is transferred into the neural network up to a certain level. While conflicting gradients may occur for enforcing various types of boundary conditions,  governing equations and other physical requirement such as symmetry and thermodynamic constraints. As such, providing a robust training strategy to handle the multiple objectives and tasks is a crucial step to reduce the manual trial-and-error and improve the practicality of PINN.

\subsection{Organization of this paper}
The rest of the paper is organized as follows. We first introduce a variety of tasks and objectives that are used to constitute the loss function and the pre-training step that involves labeled data (Section \ref{sec:lossfunction}). This section is followed by the strategies we outlined to solve the multi-task PINN problems in Section \ref{sec:grad-surg}. In particular, the gradient surgery that overcomes the conflicting gradient issue. In Section \ref{sec:net2net}, the Net2Net technique is adopted to improve the efficiency of the training of large neural network that leads to high-resolution solution. Particularly, it introduces a way to initialize large neural network in a more stable manner.
 Numerical examples are then used to tested the versatility and robustness of the training procedure and compared with benchmark results (Section \ref{sec:numric}),  followed by concluding remarks that summarize the major findings. 

\subsection{Notations}
 As for notations and symbols, bold-faced letters denote tensors (including vectors which are rank-one tensors); 
the symbol '$\cdot$' denotes a single contraction of adjacent indices of two tensors
(e.g.\ $\vec{a} \cdot \vec{b} = a_{i}b_{i}$ or $\tensor{c} \cdot \vec{d} = c_{ij}d_{jk}$ );
the symbol `:' denotes a double contraction of adjacent indices of tensor of rank two or higher
(e.g.\ $\tensor{C} : \vec{\varepsilon^{e}}$ = $C_{ijkl} \varepsilon_{kl}^{e}$);
the symbol `$\otimes$' denotes a juxtaposition of two vectors
(e.g.\ $\vec{a} \otimes \vec{b} = a_{i}b_{j}$)
or two symmetric second-order tensors
(e.g.\ $(\vec{\alpha} \otimes \vec{\beta})_{ijkl} = \alpha_{ij}\beta_{kl}$). Symmetric gradient operator of a rank one tensor $\vec{u}(\vec{x})$ is defined as $\nabla^{\text{sym}}_{\vec{x}} \vec{u} = \frac{1}{2} ( \nabla_{\vec{x}} \vec{u} + \nabla_{\vec{x}}^T \vec{u} )$ where superscript $T$ declares the transpose operator.

\remark{In the rest of the paper, when we use any finite element solution, we solve the corresponding problem by the standard Bubnov-Galerkin method with linear elements and single quadrature point rule. The FEM formulation is implemented in the FEniCS project \citet{alnaes2015fenics}. Since there is no direct access to the derived fields at the quadrature points in the FEniCS solver, we use the quadrature space (a space with no coupling between elements) to indirectly project primal or derived fields to the quadrature elements \citep{logg2012automated, bleyer2018numericaltours}.}

\remark{In this work, we implement our proposed methods and algorithms for solving forward and inverse partial differential equations via neural network approximation in PyTorch open source machine learning library \citet{paszke2019pytorch}. }

\section{Feasible physical constraints in PINN} \label{sec:lossfunction}
In this section, we provide a brief review on the formulation of PINN as a multi-objective optimization problem where 
a subset of physical constraints is used to determine the solutions. Our focus here is limited to the collocation method where a  point cloud is selected to sample how well these physical constraints are fulfilled. The formal definition of a boundary value problem can therefore be viewed as a neural network training problem in which the neuron weights are adjusted such that the discrepancy measured by a set of norms or product norms are minimized.

\subsection{Physics constraints in a plain PINN}
Within the collocation PINN framework \citep{dissanayake1994neural,lagaris1998artificial,sirignano2018dgm,raissi2018deep,raissi2019physics}, the solution field (e.g., temperature $T$ or displacement $\vec{u}$) is approximated by a feed-forward multi-layer neural network parametrized with unknowns $\vec{\theta}$; 
for simplicity, biases and weights of the neural network are concatenated in the vector $\vec{\theta} \in \mathbb{R}^{N_{\text{dof}_\text{nn}}}$ where $N_{\text{dof}_\text{nn}}$ is the total number of learnable parameters (unknowns) of the neural network. 
The abbreviations \emph{nn} and \emph{dof} stand for neural network and degree of freedom, respectively.
The unknown parameters are found by solving an optimization problem that minimizes the errors, measured by an appropriate norm (usually mean square), for satisfying, for instance, the following constraints: 
\begin{itemize}
\item PDE residual $\mathcal{L}_{\text{pde}}(\vec{\theta})$ at the chosen collocation points defined in domain $\{ \vec{x}_i^\tpde \}_{i=1}^{N_\tpde} \in \Omega$
\item  Dirichlet boundary conditions $\mathcal{L}_{\text{dbc}}(\vec{\theta})$ over Dirichlet boundary surface $\{ \vec{x}_i^\tdbc \}_{i=1}^{N_\tdbc} \subset \partial \Omega$,
\item  Neumann boundary conditions $\mathcal{L}_{\text{nbc}}(\vec{\theta})$ over Neumann boundary surface $\{ \vec{x}_i^\tnbc \}_{i=1}^{N_\tnbc} \subset \partial \Omega$. 
\end{itemize} 
The abbreviations \emph{pde}, \emph{dbc}, and \emph{nbc} stand for partial differential equations, Dirichlet boundary condition, and Neumann boundary condition, respectively. 
At this point, for the sake of generality, we do not provide the exact mathematical expression for each loss term since it depends on the PDE form, and later, we will precisely define these loss terms based on the application. In the simplest form, the plain PINN loss function includes the following three terms:
\begin{equation}
\mathcal{L}_{\text{tot}}(\vec{\theta})  = 
\mathcal{L}_{\text{pde}}(\vec{\theta}) + \mathcal{L}_{\text{dbc}}(\vec{\theta}) + \mathcal{L}_{\text{nbc}}(\vec{\theta}) = \sum_{i=1}^{3} \mathcal{L}_{i}(\vec{\theta}), 
\label{eq:loss-plain}
\end{equation}
where $\mathcal{L}_{\text{tot}}(\vec{\theta})$ denotes the scalar-valued loss function. The specific mathematical expressions of $\mathcal{L}_{\text{pde}}(\vec{\theta})$, $\mathcal{L}_{\text{dbc}}(\vec{\theta})$, $\mathcal{L}_{\text{nbc}}(\vec{\theta})$ for different boundary value problems will be provided in Section \ref{sec:numric}.

\subsection{Auxiliary supervised learning tasks for neuron weight initialization}
Here, we propose a multi-step transfer learning strategy to systematically improve the quality of the predictions. 
Inspired by the multi-grid method, our starting point here is a coarse solution obtained either from 
a PDE solver (e.g., finite element, finite difference/volume) that we reuse as auxiliary pre-training labels
to convert the PINN learning problems into a potentially easier supervised learning from 
a point cloud with labeled data.

Assume that there is some auxiliary information at some points in the domain. 
The auxiliary information is defined as labeled data that provides some rough information about the solution field. 
For example, a coarse finite element solution at the nodal and/or quadrature points can be considered as auxiliary information.
Other possibilities include spatial data derived from digital image correlation or images. 
This additional knowledge can be easily included in the total loss function to guide the training process by amending the loss function 
in Eq. \ref{eq:loss-plain}, i.e., 
\begin{equation}
\bar{\mathcal{L}}_{\text{tot}}(\vec{\theta}) = 
\mathcal{L}_{\text{tot}}(\vec{\theta}) + 
\sum_{i=1}^{N_{\taux}} \eta^{\taux}_i \mathcal{L}_{i}^{\taux}
\label{eq:loss-aux}
\end{equation}
where $\mathcal{L}_{i}^{\taux}$ is the loss term corresponding to the $i^{\text{th}}$ auxiliary task, and $\eta^{\taux}_i \in [0, 1]$ controls the contribution of each auxiliary task. 
Since the auxiliary information is not exact, minimizer Eq. \ref{eq:loss-aux} is far away from any true optimal point of Eq. \ref{eq:loss-plain}; let's assume finding such an optimal point is feasible. 
However, intuitively, these labeled data may give some hint about the true solution. 
Therefore, we are interested in an adaptive optimization problem where initially the auxiliary information is incorporated into a supervised learning task. Once the solution is \emph{sufficiently} close to complete the supervised learning task, the corresponding objective is switch off by setting $\eta^{\taux}_i =0$. 

Gradient descent algorithm is one of the flexible algorithms that enables us to eliminate the effect of auxiliary tasks by online adjustment of parameters $\eta^{\taux}_i$ during the iterations. 
We propose two possible strategies for adjusting $\eta^{\taux}_i$: (1) its value is 1 from the beginning of the training until iteration (epoch) $t^{\taux}_i$, then its value is reduced by half every $\Delta t^{\taux}_i$ until $\eta^{\taux}_i \to 0$. (2) its value is set to zero right after $t^{\taux}_i$. These parameters can be considered as extra hyperparameters that should be fine-tuned in order to obtain the best performance.

\subsubsection{Demonstrative experiments with comparison between plain PINN and adaptively guided PINN}
\begin{figure}[h]
 \centering
 \subfigure[PINN]
{\includegraphics[width=0.4\textwidth]{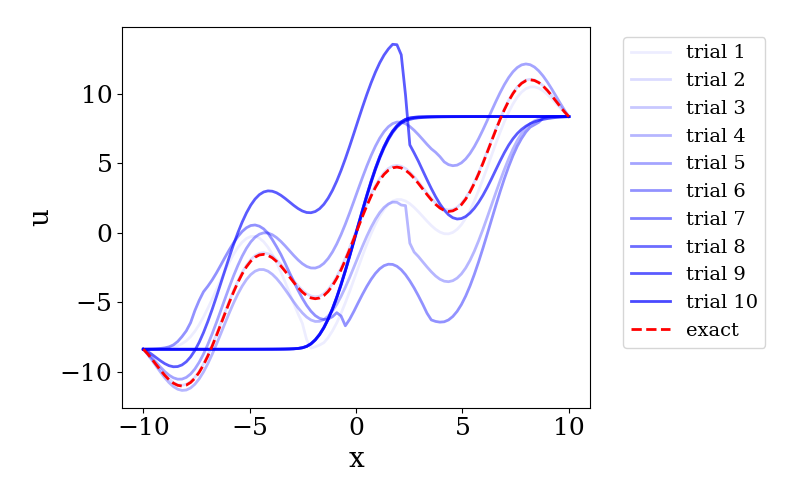}}
\hspace{0.01\textwidth}
 \subfigure[guided PINN by FEM solution]
{\includegraphics[width=0.4\textwidth]{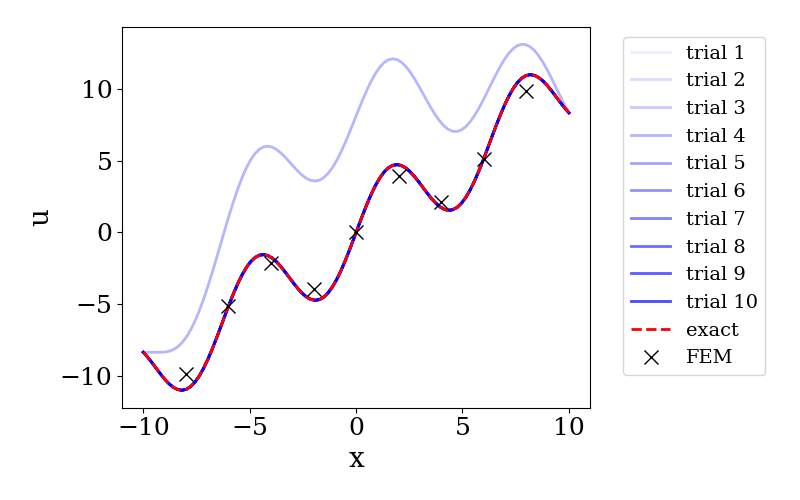}}
\hspace{0.01\textwidth}
 \subfigure[guided PINN by noisy FEM solution]
{\includegraphics[width=0.4\textwidth]{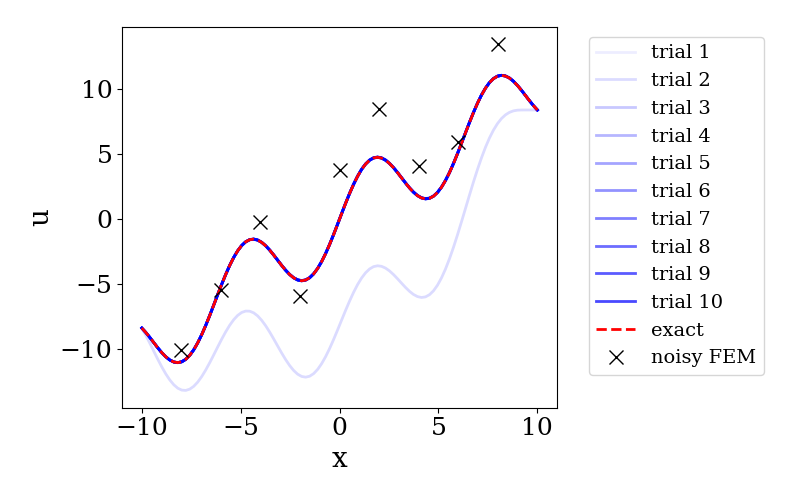}}
\hspace{0.01\textwidth}
 \subfigure[guided PINN by zeros]
{\includegraphics[width=0.4\textwidth]{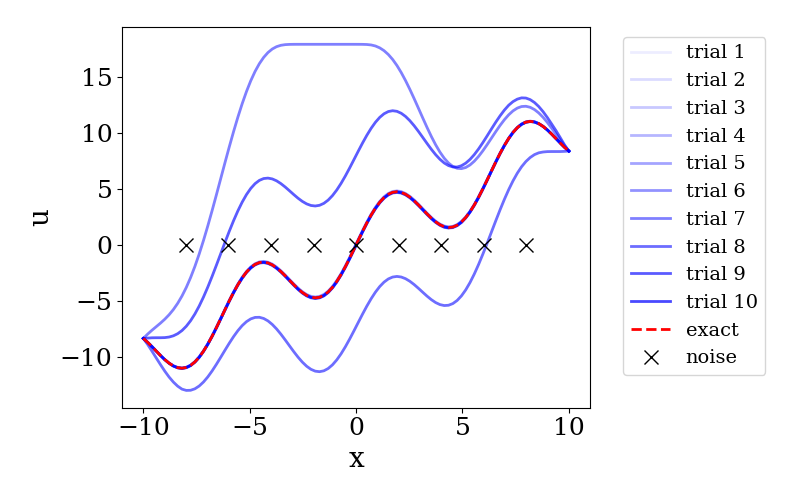}}
\hspace{0.01\textwidth}
  \caption{(a) naive PINN solution and (b) PINN with adaptive FEM guidance for different random initialization of neural network parameters. The same neural network with the same hyperparameters is used in all trials.  In (b), the FEM solution (shown by cross markers) is obtained with 10 linear elements. (c) reports the results when white Gaussian noise with standard deviation 2 is added to the FEM solution in (b). Although the FEM solution is relatively coarse or even noisy, the adaptive FEM guidance increases the robustness of the PINN solver. Appendix \ref{appx:toy1} provides details of each loss term specified for this problem.
   \label{fig::1d-sol}}
\end{figure}

To assess the effect of the auxiliary supervised learning on both the training and the robustness of the trained solution, we solve a simple 1D Poisson's equation with the following manufactured solution in Eq. \ref{eq:1dsolution}:
\begin{equation}
u_{\text{exact}} = x + \sin(x); \quad -10 \le x \le 10.
\label{eq:1dsolution}
\end{equation}
 Three PINN guided by differently labeled data are trained 10 times with different neuron weight initialization, and the results are compared with the benchmark PINN trained without labeled data. 
The only source of auxiliary information is the temperature data at nine equidistant points $\{ x^{\taux}_i \}_{i=1}^9 \in (-10, 10)$ which are shown by cross markers in Fig. \ref{fig::1d-sol}. In Fig. \ref{fig::1d-sol}(b), the auxiliary data is obtained from a coarse FEM solution with 10 elements. In Fig. \ref{fig::1d-sol}(c), the auxiliary data is a noisy version of the FEM solution. The auxiliary data in Fig. \ref{fig::1d-sol}(d) is set to zero. The details on the loss terms used in the plain PINN and PINN guided by auxiliary information are provided in Appendix \ref{appx:toy1}.

The neural network hyperparameters are set the same for all cases in Fig. \ref{fig::1d-sol}: 
multilayer perceptron with three hidden layers and 10 units in each layer is used and ADAM optimizer \citet{kingma2014adam} with initial learning rate 5e-4 is chosen. 
Each trial is a random seed for the network initialization. 
We set $t^{\taux} = 5000$ and $\Delta t^{\taux}=200$ for cases in Fig. \ref{fig::1d-sol}(b-d). 

\begin{figure}[h]
 \centering
 \subfigure[PINN]
{\includegraphics[width=0.35\textwidth]{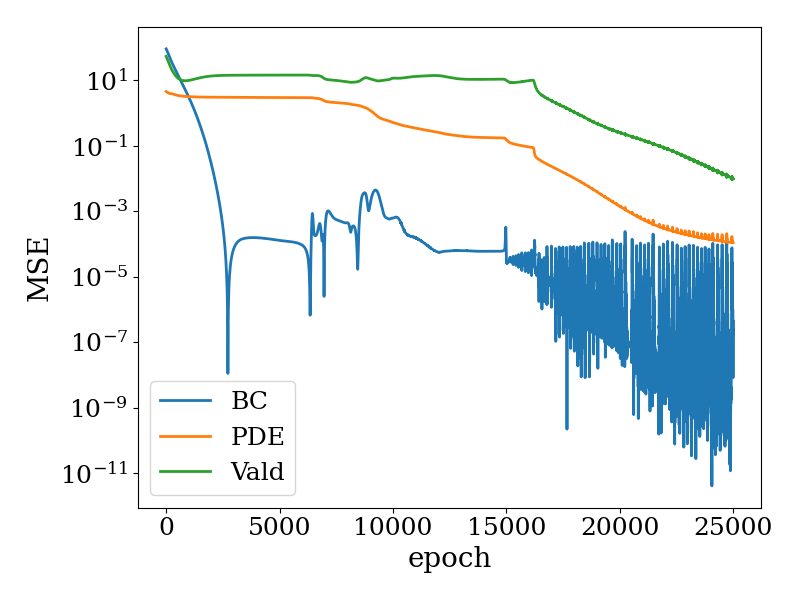}}
\hspace{0.01\textwidth}
 \subfigure[PINN+guidance by FEM]
{\includegraphics[width=0.35\textwidth]{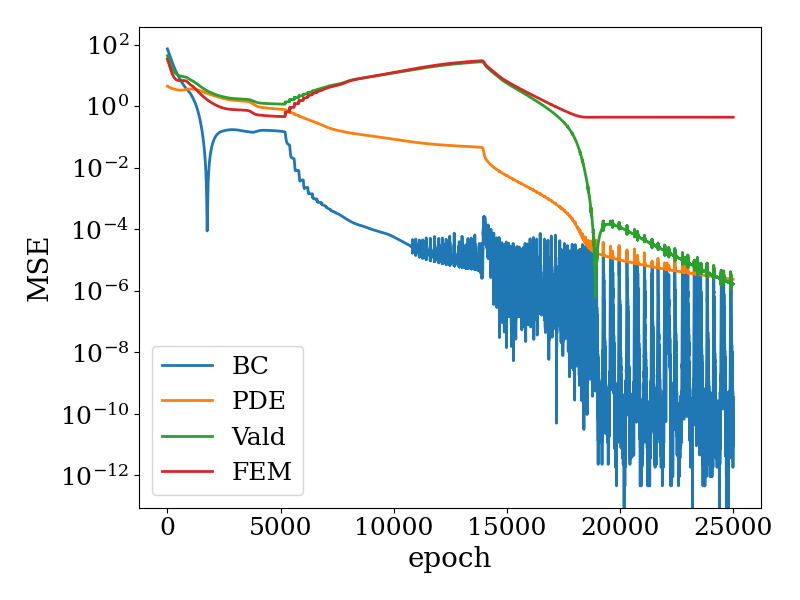}}
\hspace{0.01\textwidth}
 \subfigure[PINN+guidance by noisy FEM]
{\includegraphics[width=0.35\textwidth]{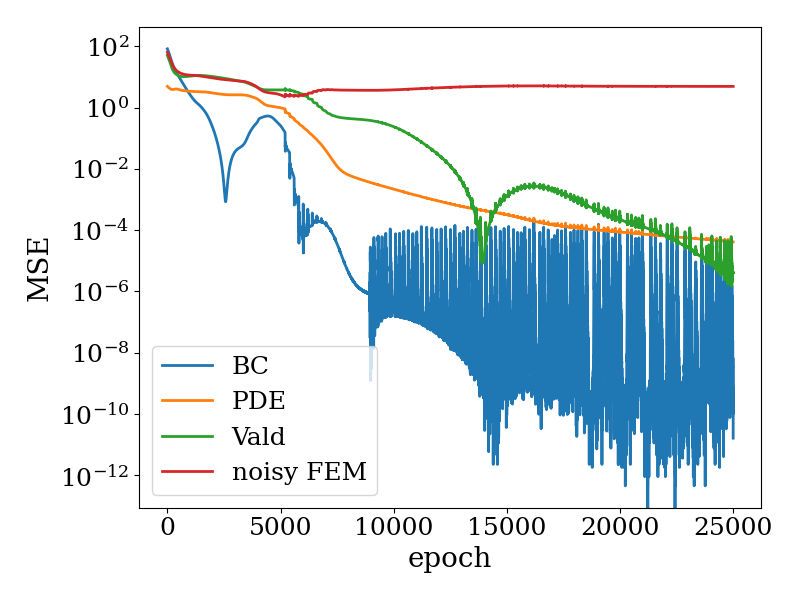}}
\hspace{0.01\textwidth}
 \subfigure[PINN+guidance by zeros]
{\includegraphics[width=0.35\textwidth]{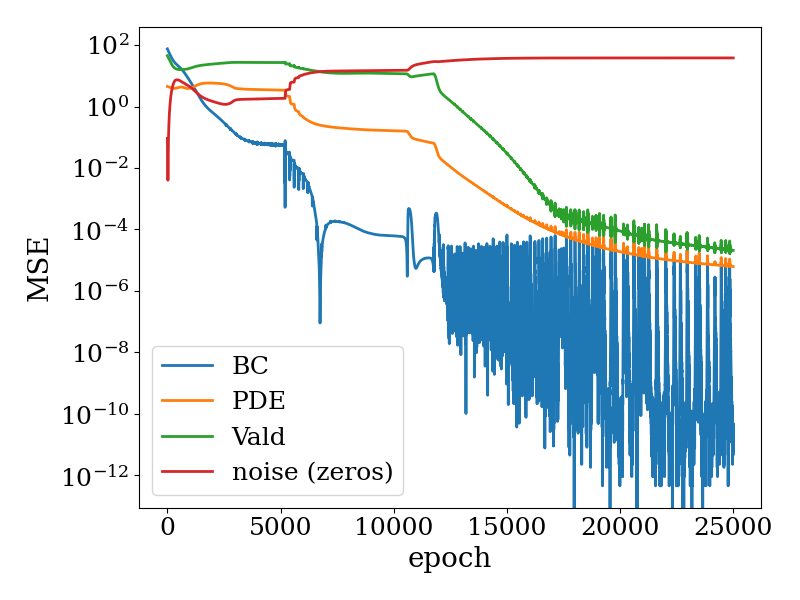}}
\hspace{0.01\textwidth}
  \caption{Training performance of PINN approach (a) without any guidance, (b) with FEM guidance, (c) with noisy FEM guidance, (d) with zero-valued guidance. All plots are shown for the trial with the lowest validation error (among 10 trials described in Fig. \ref{fig::1d-sol}) at the last epoch. MSE, BC, PDE, and FEM stand for mean square error, boundary conditions, partial differential equation, and finite element method. Validation curves, shown by Vald legend and green color, depict the discrepancy between neural network solution and exact solution at 50 random points inside the domain which are not used in training iterations. 
   \label{fig::1d-loss}}
\end{figure}

The results of this empirical study shown in Fig. \ref{fig::1d-sol} indicate that the introduced transfer learning may increase the accuracy and robustness of the PINN solver. As the results suggest the failure rate of the ADAM optimizer is considerably reduced when the pre-training method is used (see Fig. \ref{fig::1d-sol}(b-c)), even with a coarse and noisy auxiliary solution in Fig. \ref{fig::1d-sol}(c). We hypothesize that the auxiliary information helps the optimizer navigates the loss landscape in a more effective way that finds the near-optimal point faster.
For a better comparison, the training performance for the best trial of each of the four approaches is shown in Fig. \ref{fig::1d-loss}. The validation mean square error, shown by the green line in Fig. \ref{fig::1d-loss}, is defined as follows:
\begin{equation}
	\mathcal{L}_{\text{vald}} = \frac{1}{N_{\text{vald}}} \sum_{i=1}^{N_{\text{vald}}} \left(  u^{\tnn}(x_i^{\text{vald}}; \vec{\theta}) - u_{\text{exact}}(x_i^{\text{vald}})  \right)^2,
\end{equation}
where $\{ x_i^{\text{vald}} \}_{i=1}^{N_{\text{vald}}}$ are validation points which are not used in the neural network training.

The validation mean square error in cases Fig. \ref{fig::1d-loss}(b-c) is almost three orders of magnitude better than the plain PINN method in Fig. \ref{fig::1d-loss}(a).

\remark{The result in Fig. \ref{fig::1d-sol}(a) shows that some appropriate initial values for the network's weights and biases lead to a better PINN solution, provided the fixed number of iterations. Therefore, it could be helpful to have a better trial-error procedure for finding the best combination of initial weights and biases. Along this line, we will introduce an efficient algorithmic way of network initialization in Sec. \ref{sec:NN2NN}.}

\remark{In this simple problem, we use temperature data as the only source of auxiliary information. However, there could be other types of auxiliary information such as temperature gradient, heat flux, etc. These kinds of higher-order information that controls the gradient of the unknown fields can be easily incorporated into the loss function (e.g., see Sec. \ref{sec:ex-hole}). }


\remark{
It is well known that an appropriately scaled, non-dimensionalized, loss function may facilitate the gradient descent optimizer \citet{goodfellow2016deep}. However, this cannot be easily achieved in the plain collocation PINN setup since first the solution field is not known in advance, and second statistics of the PDE residual cannot be estimated even with the available field solution. 
Therefore, the coarse solution may have another advantage to provide some useful statistics such as minimum, maximum, mean, and standard deviation for an appropriate normalization of the network output or some terms in the loss function. For numerical experiments in Sec. \ref{sec:numric}, when the coarse solution is incorporated, we normalize the network output to have zero mean and unit standard deviation, and the corresponding terms in the loss function are also scaled by the inverse of the standard deviation.
}

\subsection{Other feasible objectives and metrics for correctness}
The number of primal and auxiliary objectives could be even further based on the problem formulation and physical constraints. 
For example, in the mixed formulation \citep{zhu2019physics,lyu2020mim,zhu2021local}, constitutive law (Eqs. \ref{eq:poission-const-law} and \ref{eq:elasticity-const-law}) or compatibility relations (Eq. \ref{eq:elasticity-compability}) should be included into the loss function as extra soft constraints. 
Also, there could be additional physical constraints that must be considered in the loss function to obtain an admissible response. 
For example, in the inverse problem, the material parameters should not violate thermodynamics constraints or well-posedness conditions. Moreover, there could be prior knowledge about the symmetry group of the target solution such as translation and rotational symmetries that including this knowledge as extra soft constraints may facilitate the optimization problem.

In a nutshell, the PINN problem can be considered as a multi-objective multi-task problem because it may include several different objectives for different tasks. In the next section, we discuss the challenges regarding the multi-objective problems that may arise in the PINN framework and propose an effective method to make the optimization scheme more robust for the PINN application.

\remark{The mixed formulation could have several advantages as follows. It reduces the order of regularity from $C^2$ to $C^1$ which is desirable for problems that aim to capture discontinuities. Also, it may reduce the overall computational cost since the order of partial derivatives scales exponentially in terms of time complexity in the automatic differentiation algorithm \citet{zhu2021local}.}

\section{Gradient surgery for handling multiple physics constraints}
\label{sec:grad-surg}
The multiple objectives that are used to measure the correctness of the solution may lead to conflicting gradients. The consequence 
of the conflicting gradient is that it may lead to a comprised solution that is not necessarily Pareto optima
 \citep{censor1977pareto, yu2020gradient}. To overcome this conflicting gradient and the difficulty to balance different objectives, we implement the gradient surgery approach, first introduced by \citet{yu2020gradient}, for the PINN framework.

Recall that the PINN framework aims to minimize a multi-objective loss function that depends on an unknown vector $\vec{\theta}$ which parametrizes the solution of a boundary value problem:
\begin{equation}
\vec{\theta} = \argmin_{\vec{\theta}} \mathcal{L}_{\text{tot}}(\vec{\theta}),
\end{equation}
where the total loss function includes $N_{\text{obj}}$ number of objectives as $\mathcal{L}_{\text{tot}} (\vec{\theta})= \sum_{i=1}^{N_{\text{obj}}}\mathcal{L}_i (\vec{\theta})$, and $\mathcal{L}_i$ is the loss function defined for the $i^{\text{th}}$ objective. The optimal solution can be obtained via the gradient descent (GD) family algorithms which are favorable due to their scalability for large scale problems. In the common form of GD-based methods, we have the following update rule,
\begin{equation}
\vec{\theta}^t = \vec{\theta}^{t-1} - \alpha_t \nabla_{\theta} \mathcal{L}_{\text{tot}}^{t-1},
\label{eq:update_GD}
\end{equation}
where $\vec{\theta}^t$ is the updated unknown vector at $t^{\text{th}}$ iteration, $\nabla_{\theta} \mathcal{L}_{\text{tot}}^{t-1}$ is the gradient of the total loss function with respect to the unknown vector at iteration $t-1$, and $\alpha_t$ is the learning rate that controls the stability of the solution during the iterations. If the learning rate is chosen appropriately, the GD would not increase the total loss function, i.e., 
\begin{equation}
\mathcal{L}_{\text{tot}}(\vec{\theta}^{t}) \le \mathcal{L}_{\text{tot}}(\vec{\theta}^{t-1}); \forall \ t>0. 
\end{equation}
However, this desirable property is insufficient to ensure that all objectives included in the total loss function are 
minimized satisfactorily. This goal may still be achievable by the GD, but it may increase the number of iterations significantly as we explain in the following. 

\begin{figure}[h]
\centering
{\includegraphics[width=0.5\textwidth]{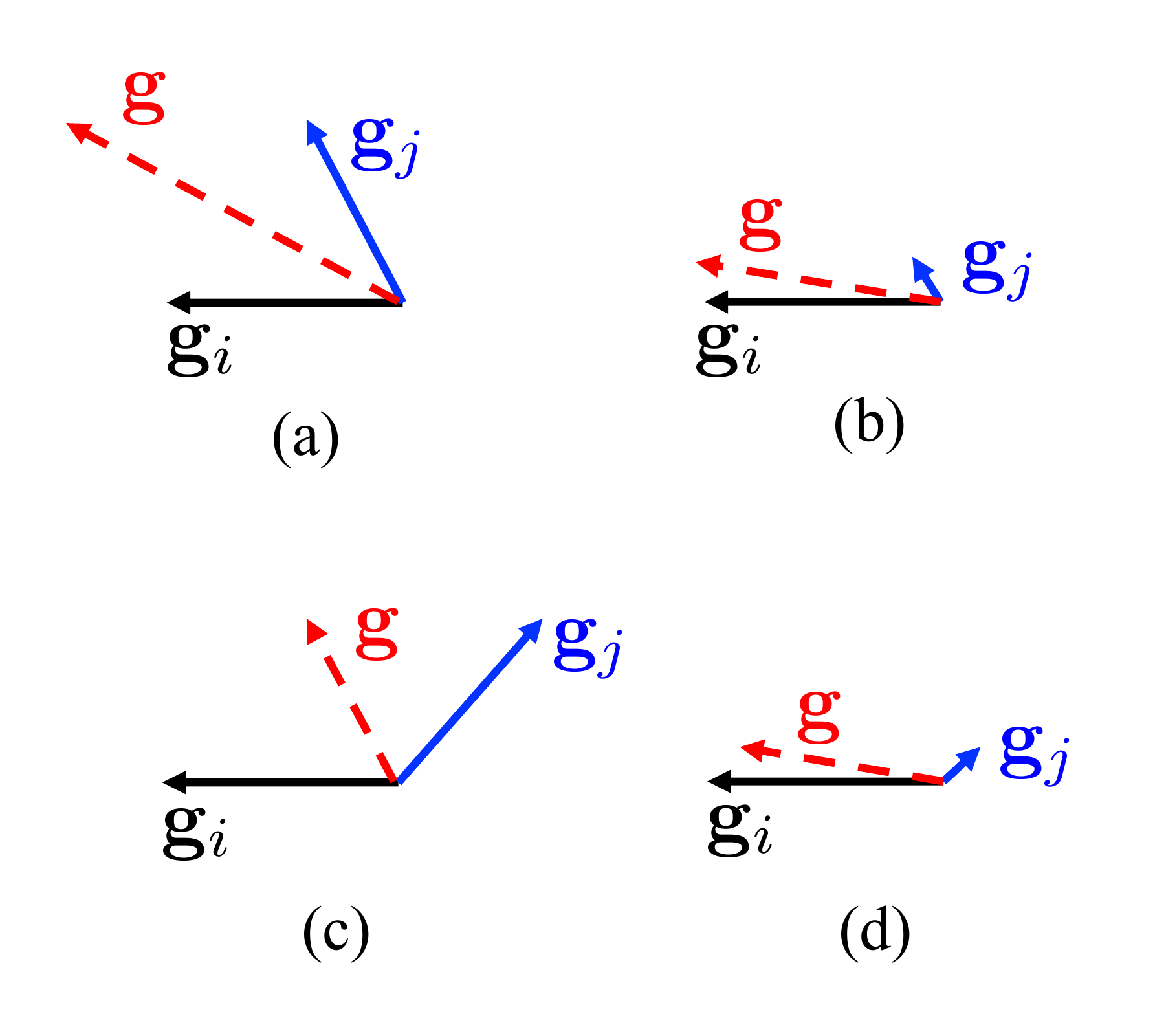}}
  \caption{ Illustration of different possibilities of the gradient descent update rule for multi-objective minimization problems. $\vec{g}$ is the gradient vector of the total loss function $\vec{g} = \vec{g}_i + \vec{g}_j$ and $\vec{g}_i$ and $\vec{g}_j$ are gradient vector of $i^{\text{th}}$ and $j^{\text{th}}$ objectives, respectively. 
  (a) there is no conflicts between vectors $\vec{g}_i$ and $\vec{g}_j$, i.e., $\vec{g}_i \cdot \vec{g}_j \ge 0$, and their norms have relatively the same scale. 
  (b) there is no conflicts between vectors $\vec{g}_i$ and $\vec{g}_j$ but their norms are not comparable, i.e., $||\vec{g}_i||_2 \gg ||\vec{g}_j||_2$. 
  (c) vectors $\vec{g}_i$ and $\vec{g}_j$ have conflicts, i.e., $\vec{g}_i \cdot \vec{g}_j < 0$, but their norms have relatively the same scale. 
  (d) vectors $\vec{g}_i$ and $\vec{g}_j$ have conflicts and their norms are not comparable. This figure is adopted and redrawn from \citet{yu2020gradient}.
   \label{fig::grad-vec-conf}}
\end{figure}

Let's define the gradient vector of each objective as $\vec{g}_i = \nabla_{\vec{\theta}} \mathcal{L}_i(\vec{\theta})$ and the total gradient vector as $\vec{g} = \nabla_{\theta} \mathcal{L}_{\text{tot}}(\vec{\theta}) = \sum_{i=1}^{N_{\text{obj}}} \vec{g}_i$. 
Geometrically, the total gradient vector $\vec{g}$ implies the descent direction that reduces the total loss. 
However, this direction may not be the descent direction for each individual sub-task objective $\mathcal{L}_i$ as illustrated in Fig. \ref{fig::grad-vec-conf}(c-d). 
For problems where accuracy of each objective have crucial impact on the accuracy of other objectives such as PINN (e.g., the solution field is highly correlated by the boundary condition, and a wrong boundary condition leads to a completely different solution), it is important to avoid scenarios depicted in Fig. \ref{fig::grad-vec-conf}(b-d). 
If at an iteration the gradient norm of one objective relatively dominates others, e.g., Fig. \ref{fig::grad-vec-conf}(b,d), then the contribution of other objectives will be relatively ignored in the the update rule Eq. \ref{eq:update_GD} which is not desirable. Also, it is not desirable to have situations where objectives have conflicts to each other ($\vec{g}_i \cdot \vec{g}_j < 0$) since in the update step their individual effect will be reduced (see Fig. \ref{fig::grad-vec-conf}(c)). 

\begin{figure}[h]
\centering
{\includegraphics[width=0.7\textwidth]{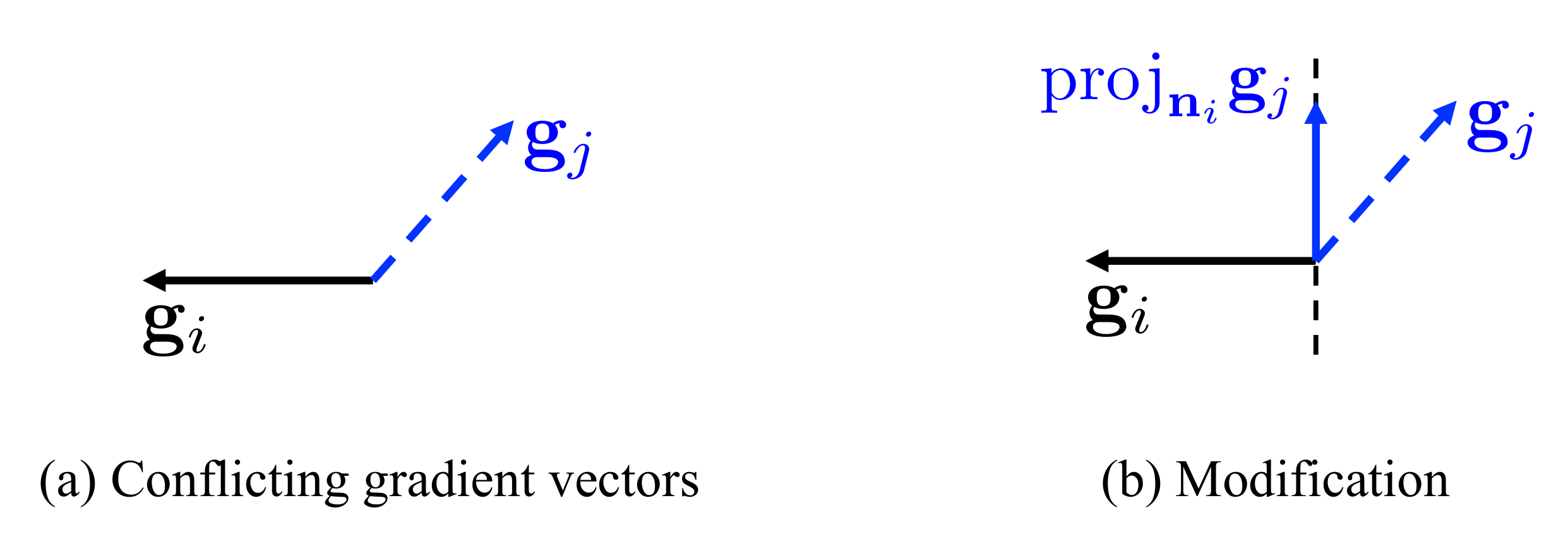}}
  \caption{Gradient surgery idea: (a) before modification where gradient vectors have conflicts. (b) one of the vectors are projected to the normal plane of the other $\text{proj}_{\vec{n}_i} \vec{g}_j = \vec{g}_j - \frac{\vec{g}_j \cdot \vec{g}_i}{\norm{\vec{g}_i}^2} \vec{g}_i$. Notice that the surgery process not only modifies the directions but also re-scales the gradient vectors. This figure is adopted and redrawn from \citet{yu2020gradient}.
   \label{fig::grad-surg-idea}}
\end{figure}

To address the above issues within the PINN framework, we use the method originally introduced by \citet{yu2020gradient} originally proposed for computer vision and reinforcement learning applications. The main idea to handle each pair of conflicting vectors is 
by projecting one of the conflicting gradient vector $\vec{g}_i$ onto the normal plane of the other gradient vector $\vec{g}_j$, and use the projection to adjust the descent direction, 
as schematically depicted in Fig \ref{fig::grad-surg-idea}. 
The overall algorithm combined by the gradient descent type optimizer is described in Alg. \ref{alg:grad-surg}, and we refer interested readers to \citet{yu2020gradient} for more information.

\begin{algorithm}[H] 
\caption{Gradient surgery algorithm}\label{alg:grad-surg}
\begin{algorithmic}[1]
\Require{$\vec{\theta}^1$}\Comment{initialize unknowns}
\Statex
    \For{$t \gets 1$ to $N_{\text{itr}}$}\Comment{$N_{\text{itr}}$ is number of gradient descent (GD) iterations}
        \State {Compute objectives and their gradients $\{\mathcal{L}_i(\vec{\theta}^t)\}_{i=1}^{N_{\text{obj}}}$, $\{\vec{g}_i = \nabla_{\vec{\theta}^t} \mathcal{L}_i\}_{i=1}^{N_{\text{obj}}}$}
        \State {Initialize modified gradients $\{\tilde{\vec{g}}_i \gets \vec{g}_i \}_{i=1}^{N_{\text{obj}}}$}
        \For{$i \gets 1$ to $N_{\text{obj}}$}
        		\For{$j \in \mathcal{D} $} \Comment{$\mathcal{D}$ is a random shuffle of indexes 1 to $N_{\text{obj}}$ excluding index $i$}
				\If{$\tilde{\vec{g}}_i \cdot \vec{g}_j < 0$}
					\State update the modified gradient $\tilde{\vec{g}}_i = \tilde{\vec{g}}_i - \frac{\tilde{\vec{g}}_i \cdot \vec{g}_j}{\norm{\vec{g}_j}^2} \vec{g}_j$
				\EndIf
         	\EndFor
         \EndFor
        \State {Aggregate gradient vectors $\tilde{\vec{g}}_{\text{tot}} = \sum_{i=1}^{N_{\text{obj}}} \tilde{\vec{g}}_i$ }
        \State {Update primal unknowns $\vec{\theta}^{t+1}$ $\gets$ {$\text{GD}(\vec{\theta}^t, \tilde{\vec{g}}_{\text{tot}})$}} \Comment{GD is a gradient descent type update rule}
    \EndFor
    \State \Return {$\vec{\theta}$}
\end{algorithmic}
\end{algorithm}

\remark{Previous research concerning multi-objective PINN \citep{raissi2019physics,lu2021deepxde,haghighat2021physics} offers a weighted aggregation of loss terms to adjust the scaling issue of each loss term and addresses the gradient norm saturation issue discussed above. However, the best weight for each loss term is not known apriori and is found by a greedy search approach. This idea has two major limitations: (1) when the number of objectives increases the computational cost of greedy search among many real values will be exponentially increased, and (2) even it could address the gradient norm saturation issue it cannot resolve the gradient conflict issue, hence it may not be able to find the Pareto-optimal solution within a reasonable amount of computational resources.}

\begin{definition}[Pareto dominant solution]
Solution $\vec{\theta}^{\text{pareto}}$ is Pareto dominant to $\vec{\theta}$ if $\mathcal{L}_i(\vec{\theta}^{\text{pareto}}) \le \mathcal{L}_i(\vec{\theta}); \forall 1 \le i \le N_{\text{obj}}$ (cf. \citet{sener2018multi}).
\end{definition}

\section{Adaptive neural network architecture via Net2Net weight initialization}\label{sec:net2net}
Previous studies \citep{glorot2010understanding,he2015delving,goodfellow2016deep, cyr2020robust} indicate that neural network weights and biases must be initialized appropriately to facilitate training of machine learning tasks such as regression and classification. 
The importance of neuron weight initialization has also been showcased in the demonstrative example 
 (see Fig. \ref{fig::1d-sol}(a)) where a plain PINN is only capable of recovering the analytical solution when 
 the network is initialized appropriately. 
 An effective way of network initialization may depend on the optimization landscape which is directly related to the type of activation function and the loss function, among other factors. 
 \citet{glorot2010understanding} and \citet{he2015delving} propose statistically consistent methods to initialize network's weights and biases. Based on the specific type of activation function used in the neural network, the bounds of each layer output can be controlled. While these approaches may significantly reduce the chance of failed training compared to an arbitrary initialization, there is no guarantee that the gradient descent will always succeed as demonstrated in the numerical experiments that involve regression and classifications. 
 Since training PINN to solve boundary value problems is by no mean an easier task, a robust and efficient approach that finds the appropriate weight and bias initialization could be instrumental for the success of the PINN training.  

In this work, we introduce a weight initialization strategy that combines the Gaussian weight initialization (Section \ref{sec:nnparam})
in \citet{glorot2010understanding} and the Net2net transfer learning (Section \ref{sec:NN2NN}) to help us 
initializing a large neural network efficiently. The key idea is to improve the efficiency of the Gaussian weight initialization 
by limiting it in a small neural network of a smaller parametric space $\vec{\theta}$, then use the Net2Net transfer learning to widen the neural network. Since the Net2net transfer learning will not alter the learned function, this 
setup will be more efficient than directly performing the Gaussian weight initialization on the wide neural network. 

\subsection{Neural network parametrization} \label{sec:nnparam}
We use a multilayer perceptron (MLP) neural network architecture $\mathcal{F}_{\vec{\theta}}(\vec{x}): \mathbb{R}^{d_\text{in}} \to \mathbb{R}^{d_{\text{out}}}$ with $L$ hidden layers that maps $d_{\text{in}}$ dimensional vector space to $d_{\text{out}}$ dimensional vector space defined as follows:
\begin{equation}
\begin{split}
&\mathcal{F}_{\vec{\theta}}( \vec{x} ) = \tensor{W}^L \vec{x}^{L} + \vec{b}^L,\\
&\vec{x}^{i+1} = \sigma(\vec{W}^i \vec{x}^i  + \vec{b}^i) \ \text{for} \  0 \le i < L,\\
&
\vec{x}^0 = \vec{x},
\end{split}
\end{equation}
where $\tensor{W}^i \in \mathbb{R}^{h_{i+1} \times h_i}$ and $\vec{b}^i \in \mathbb{R}^{h_{i+1}}$ are weight matrix and bias vector of the $i^{\text{th}}$ hidden layer, $h_{i}$ is the number of neurons in the $i^{\text{th}}$ hidden layer, and $\vec{\theta}$ concatenates all the trainable parameters $ \vec{\theta} = \{\tensor{W}^i, \vec{b}^i \}_{i=0}^{L}$. The activation function $\sigma(\vec{x}): \mathbb{R}^n \to \mathbb{R}^n$ applies a nonlinear operation on each component of an $n$ dimensional vector. 
In this work, we mostly use hyperbolic tangent activation function, i.e., $\sigma(x) = \tanh(x)$ which is also suggested by other studies \citep{raissi2019physics,lu2021deepxde,jagtap2020adaptive}, unless otherwise specified. Notice that the choice of activation function may considerably affect the performance of the PINN solver \citep{raissi2018deep,jagtap2020adaptive}.

For regression problems where the activation function of the neurons is the hyperbolic tangent function, we initialize 
the components of the weight matrix $\tensor{W}^i$  by sampling in a Gaussian distribution  with a zero mean and the standard deviation equal to $\sqrt{\frac{2}{h_i + h_{i+1}}}$. 
The initial bias vector is set to zero. 
This type of normalized initialization has been tested in \citet{glorot2010understanding}  and previously used for PINN in \citet{raissi2019physics}. The finding from \citet{glorot2010understanding}  suggests
that it helps the layer-to-layer transformations maintain magnitudes of activation and gradient for the tanh network and hence we adopted it in this paper.

We begin the training with this initialization 
for a skinny MLP and solve the PINN problem with different random seeds. Since we start with a skinny network, the computational cost for training is affordable.
Then, we select the network with the best performance and widen this network based on the Net2Net algorithm described in the next subsection. In this way, we can gradually improve the approximation, and more importantly we start training bigger network with more stable initialization.

It is worth noting that the previous theoretical guidelines for network initialization assume scaled input-output pairs. However, as mentioned earlier, the output for the PINN solver is not known in advance, and hence it is not obvious how to appropriately scale the output. This is another motivation for employing a coarse solution as an auxiliary task in the PINN framework.

\subsection{Widening neural network: Net2Net teacher-student transfer learning}  \label{sec:NN2NN}
In this section, we introduce an algorithm to transfer knowledge from a smaller neural network (teacher) to a larger one (student) via Net2Net algorithm \citet{chen2015net2net}. This algorithm aims to  \emph{exactly} preserve the function output between the two networks, i.e., $\mathcal{F}_{\vec{\theta}_{\text{teacher}}}(\vec{x}) = \mathcal{F}_{\vec{\theta}_{\text{student}}}(\vec{x})$.

\begin{figure}[h]
\centering
{\includegraphics[width=0.7\textwidth]{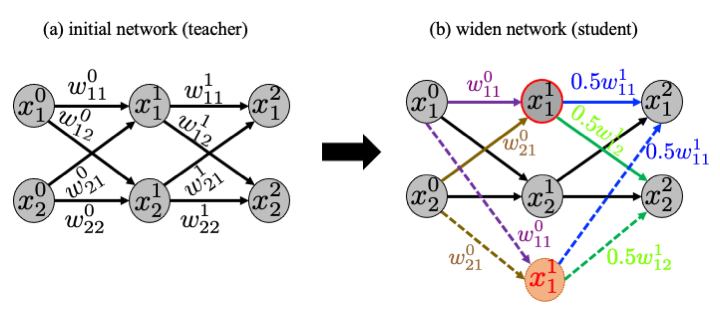}}
  \caption{(a) initial network with one hidden layer. (b) widen network by one additional hidden neuron. Net2Net aims to set new weights and biases based on the old values such that the output layer remains unchanged. This way guarantees the function output is preserved while the new network has more trainable unknowns. In this schematic representation, the contribution of biases is not included to avoid unnecessary complexity.
   \label{fig::net2net}}
\end{figure}

For the sake of argument, we assume that the teacher network has one hidden layer as shown in Fig. \ref{fig::net2net}(a) with two hidden neurons, and we aim to widen this network by one additional hidden neuron $x_3^1$. 
The strategy to preserve the last layer outputs $x_1^2$ and $x_2^2$ is to randomly set the additional hidden unit (filled by orange color in Fig. \ref{fig::net2net}(b)) to one of the teacher's hidden unit, e.g., $x_3^1 \gets x_1^1$. 
Regardless of the activation function type, if we preserve the Affine transformation from the hidden layer to the last layer (i.e., $\tensor{W}^1 \vec{x}^1 + \vec{b}^1$) the output units in the last layer will be preserved. 
This can be done, simply, by sharing the teacher's weights (solid blue and green lines) and biases with the new additional weights (dashed blue and green lines) and biases emanating from the same root ($x_1^1$). 
The additional weights and biases of the previous layer (dashed purple and brown lines) should be exactly equal to the teacher's weights and biases (solid purple and brown lines) if the destination unit has the same value ($x_1^1$). 
This idea can be recursively applied to widen a fixed depth teacher network by any arbitrary number of hidden units at any arbitrarily chosen hidden layers. 

The general algorithm is presented in Alg. \ref{alg:net2net}. In this algorithm, the function \texttt{sample$(h_i)$} randomly chooses an integer from 1 to $h_i$. Notice that the number of units in the input and output layers remain unchanged, i.e., $\Delta h_0 = \Delta h_1 = 0$.

\begin{algorithm}[H] 
\caption{Net2Net algorithm}\label{alg:net2net}
\begin{algorithmic}[1]
\Require{Teacher's weights and biases $\{ \tensor{W}^i, \vec{b}^i\}_{i=0}^{L}$, and number of additional hidden units $\{ \Delta h_i \}_{i=0}^{L+1}$} 
\Statex
    \State {set teacher-student connectors $\{ \vec{c}^l \in \mathbb{Z}^{h_l + \Delta h_l} \}_{l=0}^{L+1}$ such that
            $c_i^l = 
                \begin{cases}
                i,& \text{if } i \le h_i\\
                \texttt{sample}(h_i) 
                & h_i < i \le h_i + \Delta h_i 
                \end{cases}$
            }
    \For{$l \gets 0$ to $L$}
        \For{$i \gets 1$ to $h_{l+1} + \Delta h_{l+1}$}
            \State {find the number of shared connections $n_{\text{sh}} = | \{ m | c^l_m = c^l_i \} | $}
                \For{$j \gets 1$ to $h_l + \Delta h_l$}
                    \State {$\tilde{w}^l_{ij} = \frac{1}{n_{\text{sh}}} w^{l}_{c^l_i c^{l+1}_j}$} \Comment{student weight}
                    \State {$\tilde{b}^l_i = \frac{1}{n_{\text{sh}}} b^{l}_{c^l_i}$} \Comment{student bias}
                \EndFor
        \EndFor
    \EndFor
    \State \Return {Student's weights and biases $\{ \tilde{\tensor{W}}^i, \tilde{\vec{b}}^i\}_{i=0}^{L}$}
\end{algorithmic}
\end{algorithm}

To assess the feasibility of the Net2Net transfer learning for the PINN framework, we solve a simple 1D heat conduction problem (see Appendix \ref{appx:toy2} for more details) where the temperature $T$ and heat flux $q = \frac{dT}{dx}$ are primary unknowns and parameterized by a single three layers MLP with silu activation function $\sigma(x)=x \cdot \text{sigmoid}(x) = \frac{x}{1+\exp(-x)}$. We do not use any theoretical results to initialize this network; as there may not be any theoretical study on silu network initialization. 
The teacher network's weights are initialized by the Gaussian distribution with zero mean and standard deviation two, and its biases are set to zero. 
Each teacher network's layer has 4 hidden neurons, and we randomly initialize 50 teacher networks. 

\begin{figure}[h]
 \centering
 \subfigure[teacher: before training]
{\includegraphics[width=0.31\textwidth]{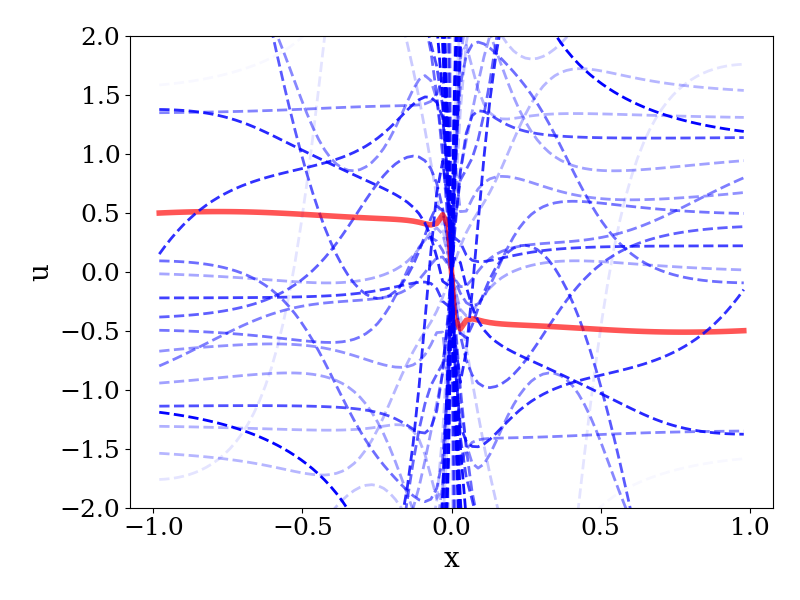}}
\hspace{0.01\textwidth}
 \subfigure[teacher: after training]
{\includegraphics[width=0.31\textwidth]{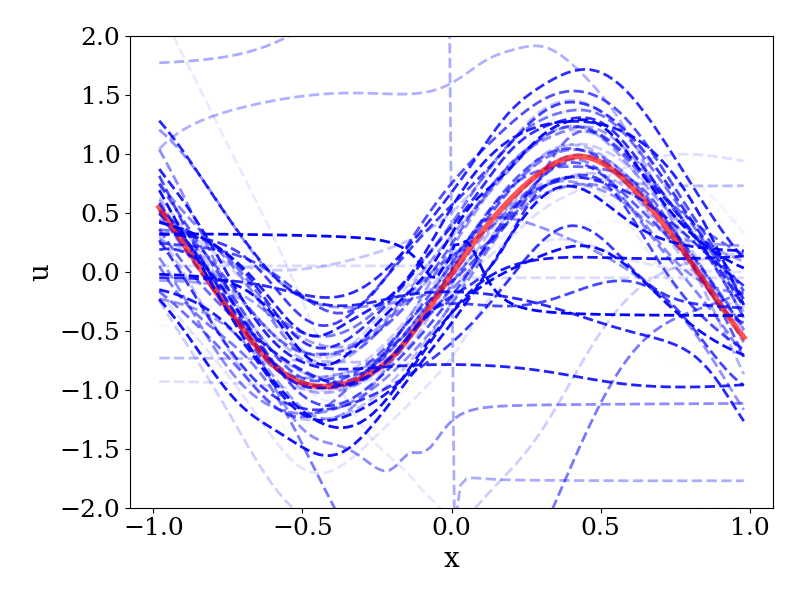}}
\hspace{0.01\textwidth}
 \subfigure[best teacher and its trained student]
{\includegraphics[width=0.31\textwidth]{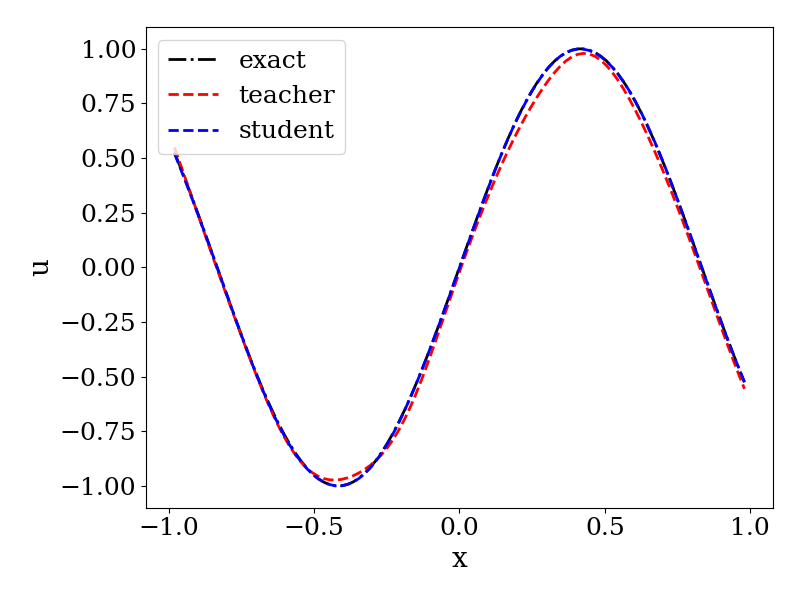}}
  \caption{Network output for $u$ (a) before and (b) after training of the teacher network. (c) comparing profile $u$ among the exact solution $u_{\text{exact}}(x) = \sin(1.2 \pi x)$, the best chosen teacher network (red-colored in Figs. (a) and (b)), and student network at the end of training. Dashed blue lines with different color transparencies indicate different neural networks initialized by different random seeds.
   \label{fig::1d-net2net-u}}
\end{figure}

\begin{figure}[h]
 \centering
 \subfigure[teacher: before training]
{\includegraphics[width=0.31\textwidth]{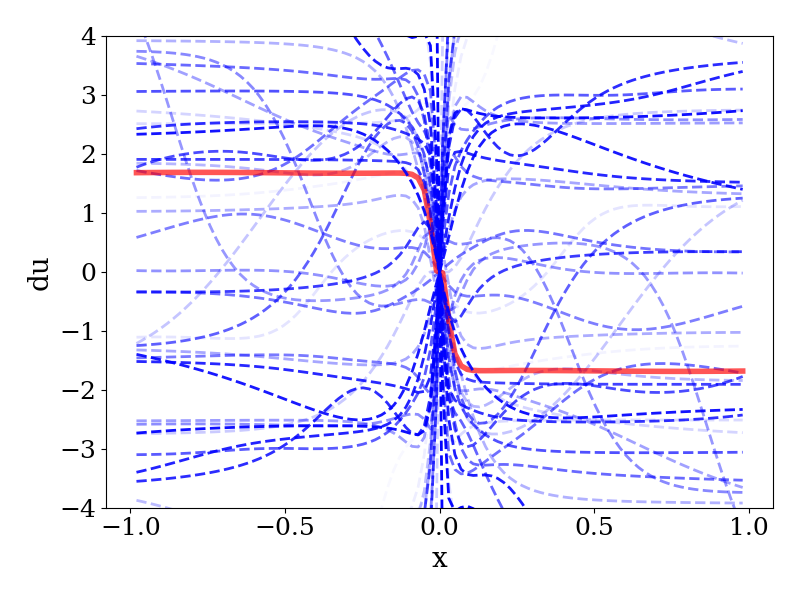}}
\hspace{0.01\textwidth}
 \subfigure[teacher: after training]
{\includegraphics[width=0.31\textwidth]{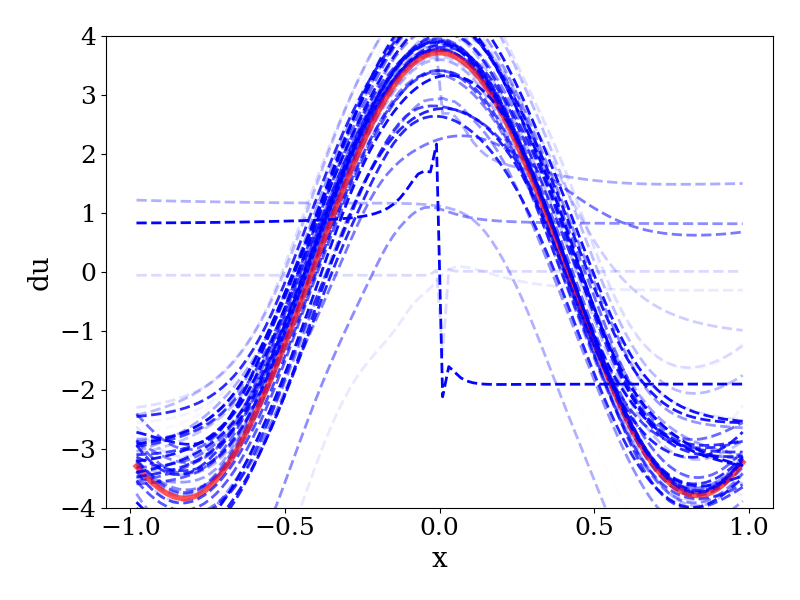}}
\hspace{0.01\textwidth}
 \subfigure[best teacher and its trained student]
{\includegraphics[width=0.31\textwidth]{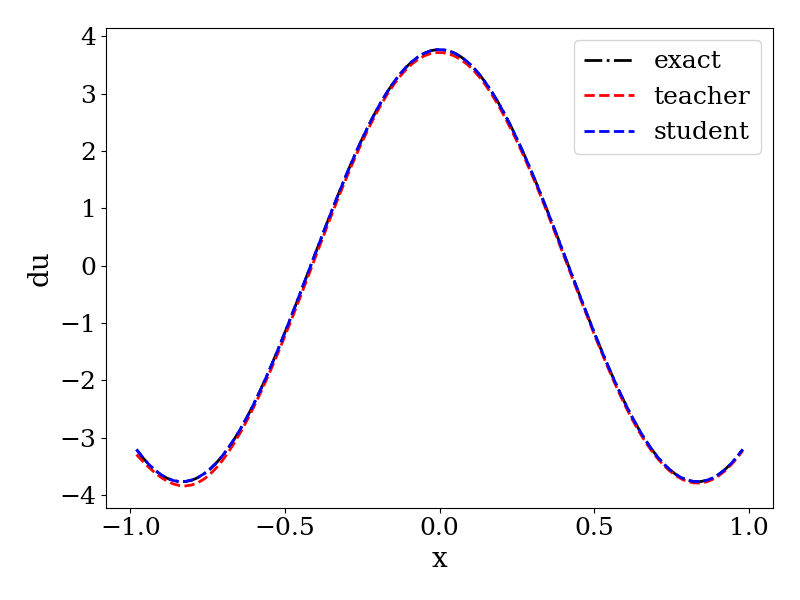}}
  \caption{Network output for $\frac{du}{dx}$ (a) before and (b) after training of the teacher network. (c) comparing profile $\frac{du}{dx}$ among the exact solution, the best-chosen teacher network (red-colored in Figs. (a) and (b)), and the student network at the end of training. Dashed blue lines with different color transparencies indicate different neural networks initialized by different random seeds.
   \label{fig::1d-net2net-du}}
\end{figure}

\begin{figure}[h]
 \centering
 \subfigure[teacher]
{\includegraphics[width=0.31\textwidth]{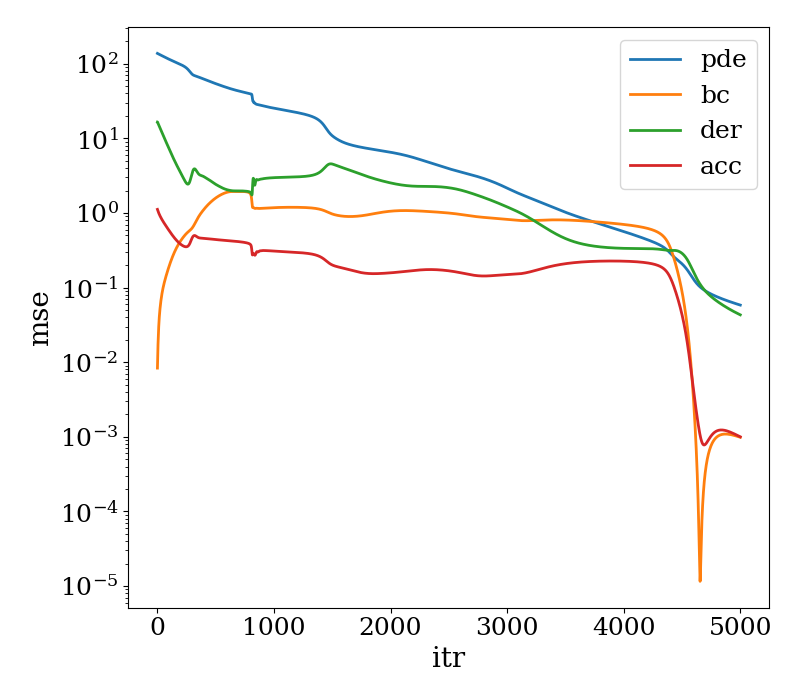}}
\hspace{0.01\textwidth}
 \subfigure[student]
{\includegraphics[width=0.31\textwidth]{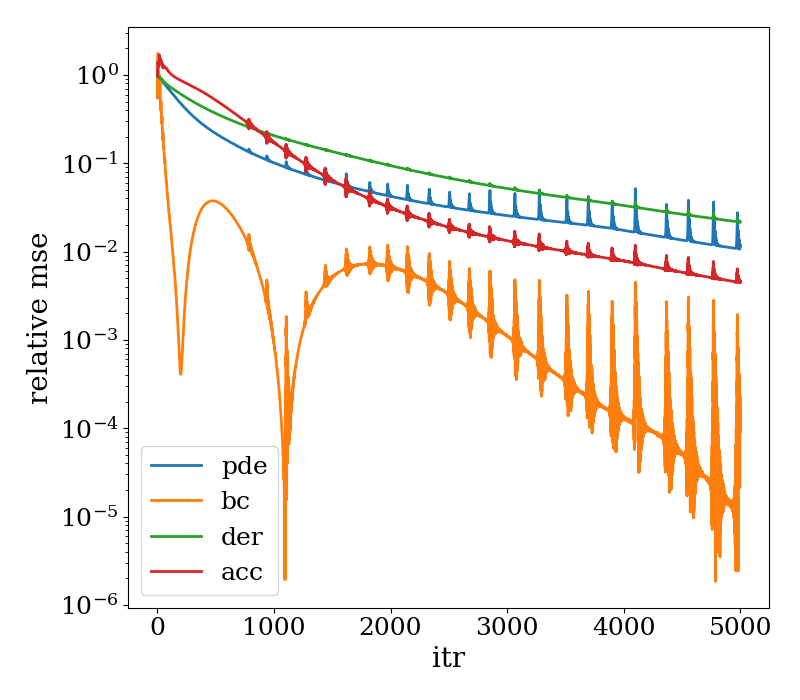}}
\hspace{0.01\textwidth}
 \subfigure[student:zoomed initial iterations]
{\includegraphics[width=0.31\textwidth]{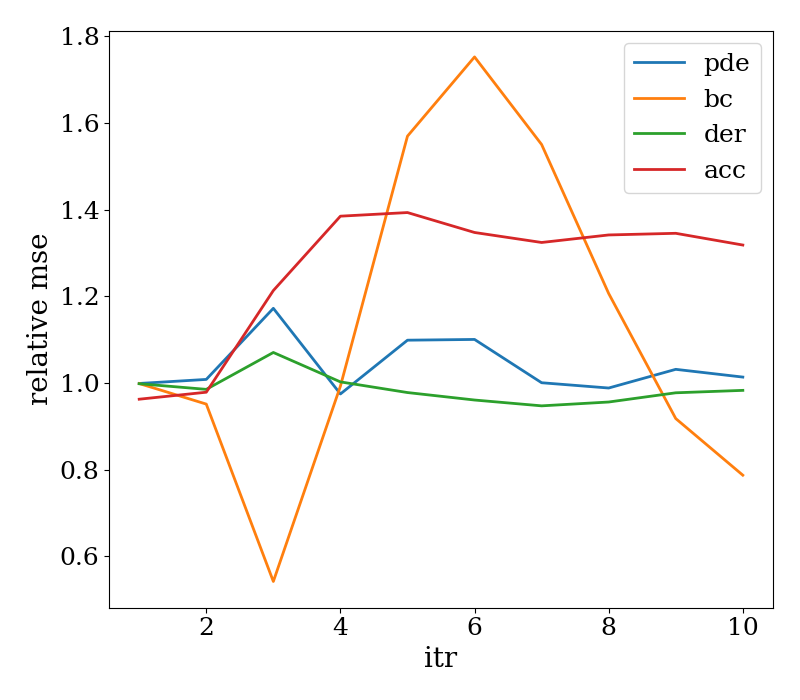}}
  \caption{ (a) mean squared errors (mse) during the training iterations (itr) of the best-chosen teacher network, corresponding to the red curves in Figs. \ref{fig::1d-net2net-u} and \ref{fig::1d-net2net-du}. (b) relative mse of the student network with respect to the last mse of the teacher network. (c) zoom area of Fig. (b) at the beginning of the training. Errors corresponding to the partial differential equations, boundary conditions, and derivative compatibility are shown by lines with legend \emph{pde}, \emph{bc}, and \emph{der}, respectively. The line with legend \emph{acc} shows the error between exact $u$ and predicted $u$ over the training collocation points.
   \label{fig::1d-net2net-loss}}
\end{figure}

Figs. \ref{fig::1d-net2net-u}(a) and \ref{fig::1d-net2net-du}(a) plot the output fields of all teacher networks before the first gradient descent iteration. 
Figs. \ref{fig::1d-net2net-u}(b) and \ref{fig::1d-net2net-du}(b) plot the output fields of teacher networks after 5000 gradient descent iterations with constant learning rate 0.001. 
As expected, initial values of network parameters considerably impact the training process, and among these 50 networks, only a few of them achieve a reasonable performance. This motivates us to not perform the trial-and-errors with big networks as it may require heavy simulations to find the near-optimal network. Instead, the computational cost for small networks is relatively low. 
Now, we select the best teacher network which is shown by red lines in Figs. \ref{fig::1d-net2net-u} and \ref{fig::1d-net2net-du}, and widen all of its hidden layers to 8 neurons. Fig. \ref{fig::1d-net2net-loss}(a) reports the training performance of the chosen teacher network. 
Fig. \ref{fig::1d-net2net-loss}(b) confirms that the student network training improves each loss term and accuracy, and it starts the learning with almost the same values of the last teacher training iteration (see Fig. \ref{fig::1d-net2net-loss}(c)). As suggested by \citet{chen2015net2net}, we reduce the learning rate for the student network by half since the solution is potentially close to the optimal, and it is widely advised to gradually reduce the learning rate when the gradient descent approaches to the optimal solution \citet{goodfellow2016deep}.

Notice that additional computational costs will be added when the gradient surgery treatment (see Sec. \ref{sec:grad-surg}) is utilized due to extra gradient calculations and conflict adjustment. Therefore, the proposed sequential learning strategy via the Net2Net algorithm could significantly reduce the computational cost by avoiding crude trial-and-errors on big networks.

\remark{Net2Net algorithm preserves only the function output and not its derivatives with respect to the input arguments. It may be more helpful to develop algorithms that also preserve the function's derivatives (higher-order) which lead to the preservation of PDE residuals between teacher and student neural networks. Notice that mixed formulations (such as the one discussed in this section) can partially address this issue since the network output includes derivatives and hence the field and it's derivative will be preserved in the Net2Net operation.}

\section{Numerical Experiments}\label{sec:numric}
In this section, we solve three 2D problems to demonstrate the potential application of proposed training algorithms for different scenarios. In the first example, the Poisson's problem is solved where the solution has a high-frequent spatial characteristic. We demonstrate that the pre-training strategy with a coarse solution is a simple but effective remedy to improve the plain PINN method
and address the training issue identified previously by  \citet{van2020optimally} (without designing special activation functions or adaptive task weighing). 
In the second example, we showcase the application of gradient surgery and pre-training strategy via a common solid mechanics benchmark problem. To facilitate the training and avoid calculation of higher-order derivatives, 
we opt to use a mixed collocation-based formulation where displacement and strain fields are parameterized with a neural network. 
This treatment nevertheless increases the number of tasks that may complicate the search for an optimized solution. 
Our comparison shows that the proposed algorithm improves the accuracy compared to those obtained from plain PINN.
In the third example, we study the application of Net2Net transfer learning and weight initialization in an inverse anisotropic Poisson's problem where gradient surgery is also utilized to handle this multi-objective optimization problem. We illustrate that the Net2Net operation can approximately preserve all the loss terms during the knowledge transformation from a teacher to a higher capacity student which has two orders of magnitude more parameters than the teacher. The performance of the teacher-student transfer learning indicates that using the smaller neural network to explore the smaller parametric space while leveraging the transfer learning to enable exploitation could be an effective strategy for training PINN.

\subsection{Two-dimensional Poisson's problem}
This boundary value problem is a benchmark case in \citet{van2020optimally} in which it has been demonstrated that 
this BVP is difficult to solve with the classical PINN approach. Here, we show that an auxiliary pre-training task enabled by the labeled data obtained from a coarse FEM simulation is sufficient to improve the PINN learned solution without any extra treatment.
We solve a 2D Poisson's problem where $\Omega \subset (0, 1)^2$ is a square domain with unit length. The source term $s(\vec{x})$ and pure Dirichlet boundary condition are set according to the exact solution:

\begin{equation}
    u_{\text{exact}}(\vec{x}) = \sin(4\pi x_2) \exp(-4\pi x_1).
\end{equation}

\begin{figure}[h]
 \centering
 \subfigure[FEM mesh]
{\includegraphics[width=0.3\textwidth]{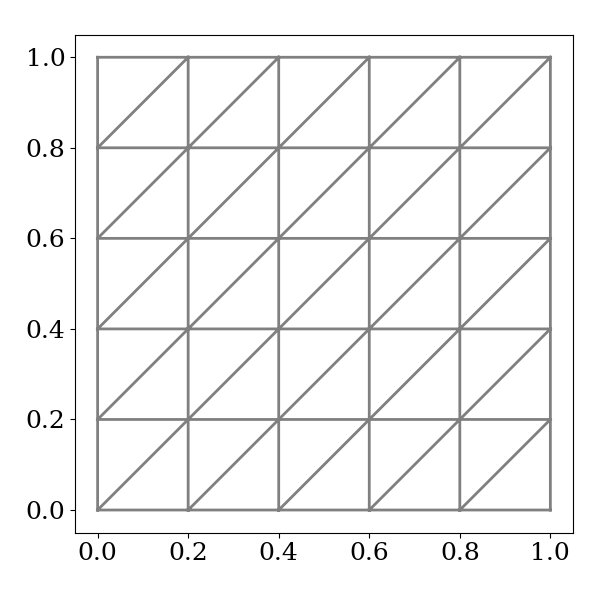}}
\hspace{0.01\textwidth}
 \subfigure[PINN training points]
{\includegraphics[width=0.3\textwidth]{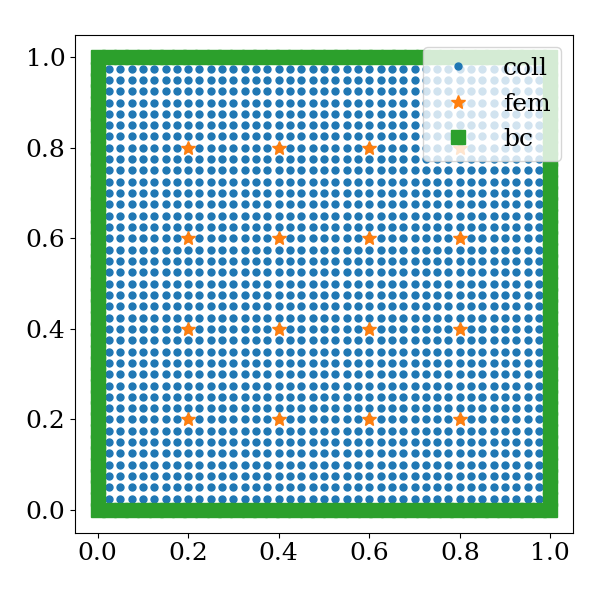}}
  \caption{ (a) coarse mesh for FEM solver. (b) domain and input points for the 2D Poisson's problem. The FEM coarse solution at orange star points in (b) is transferred into the PINN solver during the initial stages of training and eventually, this information will be eliminated. Internal collocation points (blue dot points) are chosen from a uniform $40\times40$ grid. Each side of the boundary is discretized by 40 equidistant collocation points (green square points).
   \label{fig::2d-poisson-domain}}
\end{figure}

Figure \ref{fig::2d-poisson-domain} shows the finite element mesh and PINN collocation points. The neural network has three hidden layers with the tanh activation function, and each layer has 80 hidden units. Full batch training is performed by the \textit{Adam} optimizer with a learning rate 0.001. 
Also, \textit{ReduceLROnPlateau} learning scheduler of PyTorch is used with $\textit{patience}=50$, $\textit{factor}=0.8$, and $\textit{min\_lr}=1\mathrm{e}{-6}$, for more details please refer to the explained API in the \texttt{PyTorch} website. 
The same hyper-parameters are set for all the training tasks in this problem.

\begin{figure}[h]
 \centering
 \subfigure[Exact]
{\includegraphics[width=0.3\textwidth]{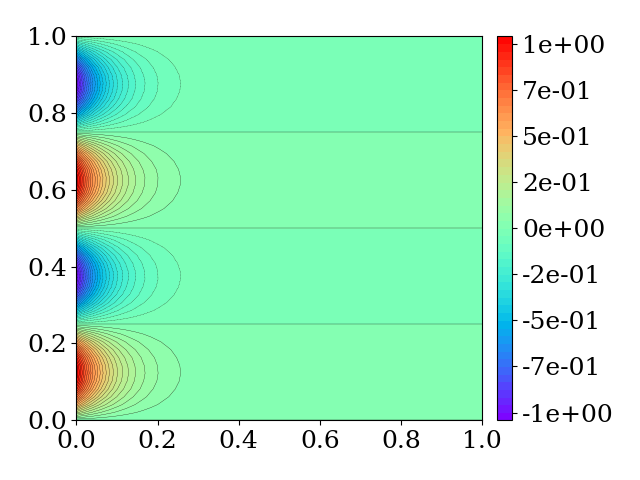}}
\hspace{0.01\textwidth}
 \subfigure[FEM+PINN]
{\includegraphics[width=0.3\textwidth]{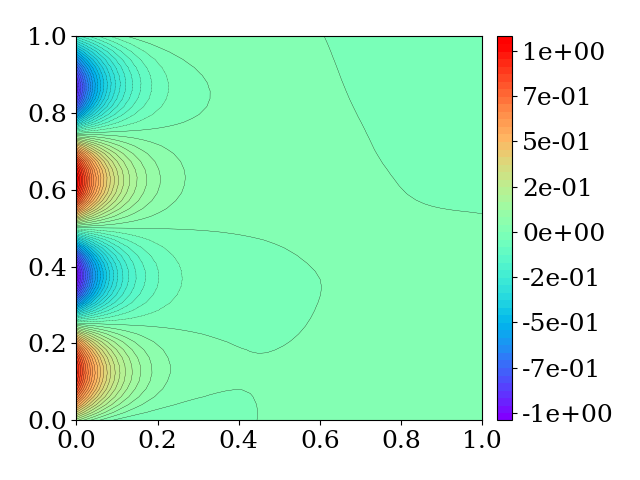}}
\hspace{0.01\textwidth}
 \subfigure[PINN]
{\includegraphics[width=0.3\textwidth]{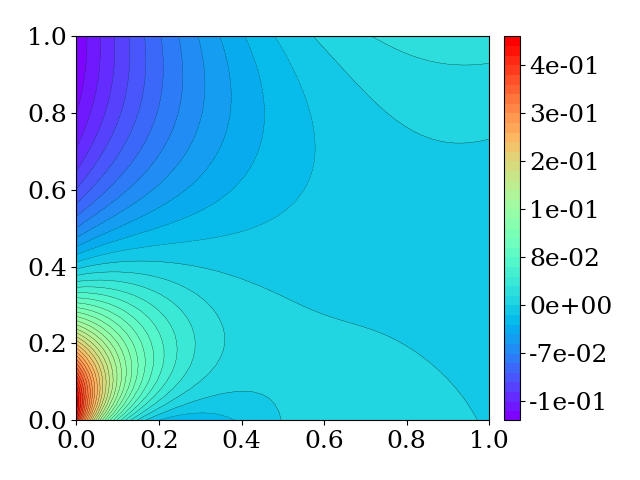}}
  \caption{ (a) analytical solution for temperature contour, (b) PINN solution guided by a coarse FEM solution, and (c) plain PINN solution. As the results show, the plain PINN with equal penalty values is not able to recover even the pattern of the solution which is also reported in \citet{van2020optimally}.
   \label{fig::2d-poisson-solution}}
\end{figure}

\begin{figure}[h]
 \centering
 \subfigure[FEM+PINN]
{\includegraphics[width=0.35\textwidth]{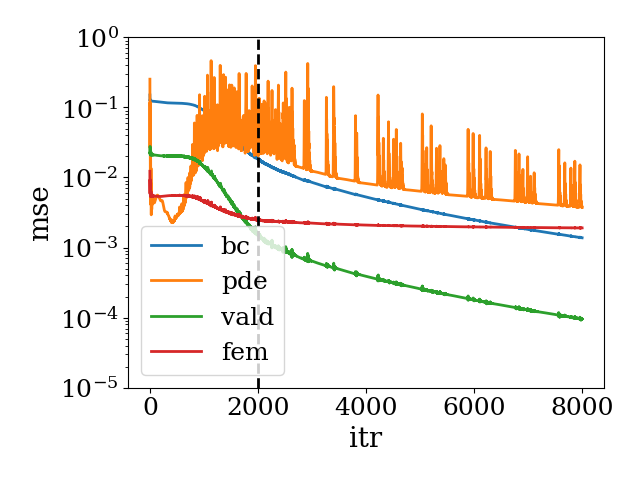}}
\hspace{0.01\textwidth}
 \subfigure[PINN]
{\includegraphics[width=0.35\textwidth]{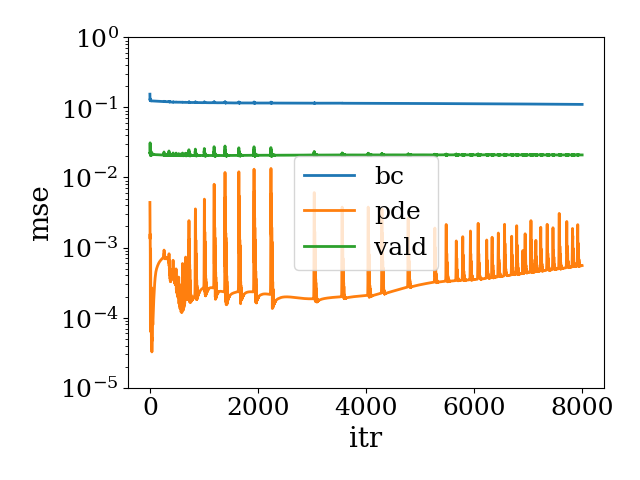}}
  \caption{ (a) mean square errors of the satisfaction of neural network solution in Dirichlet boundary condition, Poisson's partial differential equation, and finite element coarse solution during training iterations of the guided PINN approach which are shown by \emph{bc}, \emph{pde}, and \emph{fem} in the legend, respectively. (b) the same study for the plain PINN method. The penalty factor for the FEM term in the total loss function in Fig. (a) is reduced by half every 200 iterations after iteration 2000, and this is why the loss term corresponding to the FEM solution (red curve) is not improved after iteration 2000. The green curves with \emph{vald} legend show the validation mse between the exact solution and neural network prediction. The validation collocation points are 2000 randomly generated points by the Latin-hypercube sampling method and are not used in the training back-propagation. Notice that, this figure shows the training process of the final result in Fig. \ref{fig::2d-poisson-solution}.
   \label{fig::2d-poisson-loss}}
\end{figure}

Figure \ref{fig::2d-poisson-solution} compares the solution between plain PINN and the guided PINN. For the guided PINN by FEM, we do not utilize the optimization treatment (gradient surgery) explained in Sec. \ref{sec:grad-surg}. 
The results indicate that the auxiliary information from FEM can accelerate and improve the PINN training without the theoretically derived adaptive weighting introduced in \citet{van2020optimally}. The mean square errors of each term in the loss function are reported in Fig. \ref{fig::2d-poisson-loss}. As the result indicates, the guided PINN does not show the plateau behavior during the training.


\remark{The issue reported here is also related to the neural network tendency for learning solutions with low-frequency mode \citet{rahaman2019spectral}. Previous research \citep{jagtap2020adaptive,hennigh2021nvidia} develop new activation functions as one way to improve the neural network capacity for learning high frequency function. In this work, however, we propose another potential way.}

\subsection{Stress concentration around hole in infinite domain}\label{sec:ex-hole}

\begin{figure}[h]
 \centering
 \subfigure[FEM mesh]
{\includegraphics[width=0.3\textwidth]{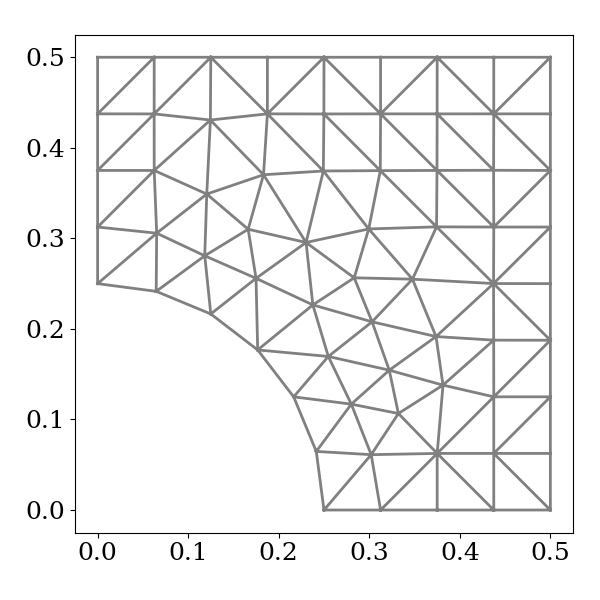}}
\hspace{0.01\textwidth}
 \subfigure[PINN training points]
{\includegraphics[width=0.3\textwidth]{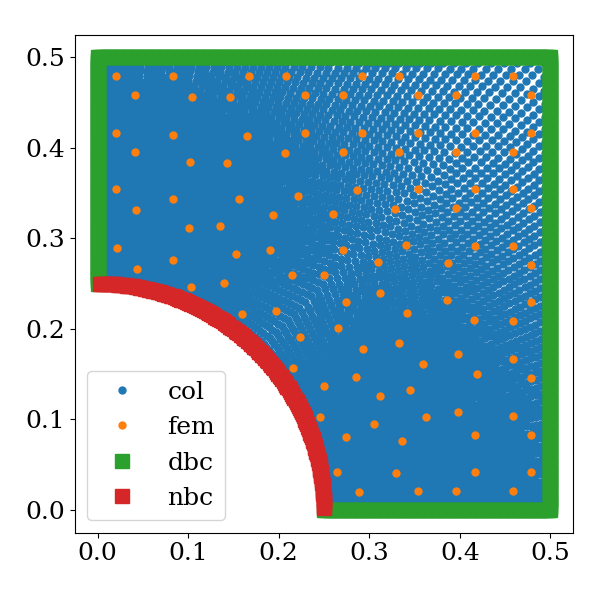}}
  \caption{ (a) coarse mesh for FEM solver. (b) domain and input points for the 2D elasticity problem. FEM points (dot orange markers) in (b) are Gaussian quadrature points of elements in (a). Square green and red markers are used for Dirichlet and Neumann boundary conditions enforcement. Collocation points for imposing PDE and strain compatibility are shown by dot blue points. Dirichlet collocation points (green square points) are chosen on each boundary side such that the distance between any two neighbors is $2.5\mathrm{e}-5$. The hole boundary has 160 equidistant Neumann collocation points (red square points). To generate internal collocation points, we choose 50 equidistant angles between $(0, \frac{\pi}{2})$ and for each angle, the ray interval between the hole boundary and side boundary is discretized by 50 points; in total 2500 internal collocation points (blue dot points).
   \label{fig::2d-solid-hole-domain}}
\end{figure}

\begin{figure}[h]
 \centering
{\includegraphics[width=0.6\textwidth]{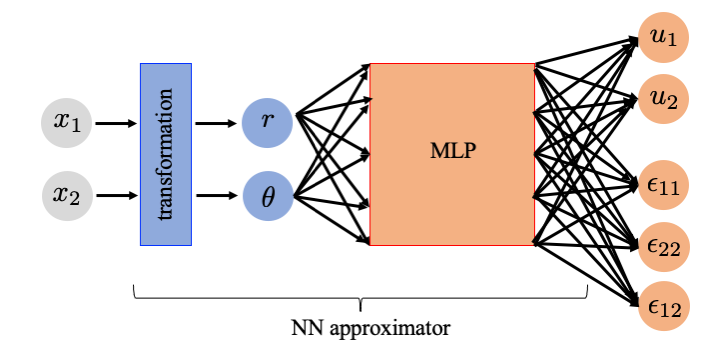}}
  \caption{ Neural network architecture that maps Cartesian coordinate as its input to displacement and strain fields. The transformation layer transforms Cartesian coordinate to polar coordinate; this choice makes the training process easier based on our experimentation. The MLP has three hidden layers with tanh activation function, and each layer has 40 hidden units.
   \label{fig::2d-solid-mixed-NN}}
\end{figure}

In this problem, we solve 2D elasticity equations for the plate with a circular hole at its center in the plane strain condition (see Fig. \ref{fig::2d-solid-hole-domain}(a)). 
This problem is formulated in the mixed form where displacement vector and strain tensor are the primary unknown fields. The neural network architecture is schematically depicted in Fig. \ref{fig::2d-solid-mixed-NN}. The loss function for the PINN solver reads as:
\begin{equation}
\begin{split}
\mathcal{L}^{\text{PINN}}_{\text{tot}}(\vec{\theta}) & = \mathcal{L}_{\text{pde}}(\vec{\theta})
+ \mathcal{L}_{\text{strn}}(\vec{\theta})
+ \mathcal{L}_{\text{dbc}}(\vec{\theta}) 
+ \mathcal{L}_{\text{nbc}}(\vec{\theta})\\
&= 
\frac{1}{N_{\text{col}}} \sum_{i=1}^{N_{\text{col}}} \norm{ \nabla_{\vec{x}} \cdot \tensor{\sigma}^{\text{NN}}(\vec{x}_i^{\text{col}}; \vec{\theta}) }_2^2
+
\frac{1}{N_{\text{col}}} \sum_{i=1}^{N_{\text{col}}} \norm{ \nabla_{\vec{x}}^{\text{sym}} \vec{u}^{\text{NN}}(\vec{x}_i^{\text{col}}: \vec{\theta})  - \tensor{\epsilon}^{\text{NN}}(\vec{x}_i^{\text{col}}; \vec{\theta})}_{F}^2\\
& \quad 
+ \frac{1}{N_{\text{dbc}}} \sum_{i=1}^{N_{\text{dbc}}} \norm{ \vec{u}^{\text{NN}}(\vec{x}_i^{\text{dbc}}; \vec{\theta}) - \bar{\vec{u}}(\vec{x}_i^{\text{dbc}})}_2^2
+\frac{1}{N_{\text{nbc}}} \sum_{i=1}^{N_{\text{nbc}}}  \norm{  \tensor{\sigma}^{\text{NN}}(\vec{x}_i^{\text{nbc}}; \vec{\theta}) . \vec{n}(\vec{x}_i^{\text{nbc}}) }_2^2
\end{split},
\label{eq:loss-pinn-solid}
\end{equation}
where the stress $\tensor{\sigma}^{\text{NN}}$ is obtained from the strain field $\epsilon^{\text{NN}}$ and $\norm{\cdot}_F$ and $\norm{\cdot}_2$ denote Frobenius and Euclidean norms, respectively. 
The mixed formulation in Eq. \ref{eq:loss-pinn-solid} allows us to bypass the need to compute the second derivatives of the displacement field. 
The prescribed displacement vector $\bar{\vec{u}}$ on the Dirichlet boundary surface is set according to the following exact solution for the infinite domain with circular imperfection of radius $r_c$ under remote traction $\sigma_0$ with Elastic modulus $E$ and Poisson's ratio $\nu$ \citet{bower2009applied}:
\begin{align}
u_{x_1} &= \frac{\sigma_0 (1+\nu) r_c}{2E}
\left( 
2(1-\nu)(\frac{r}{r_c} + \frac{2r_c}{r}) \cos \theta
+
(\frac{r_c}{r} - \frac{r_c^3}{r^3}) \cos 3\theta
\right),\\
u_{x_2} &= \frac{\sigma_0 (1+\nu) r_c}{2E}
\left( 
-
2(1-2\nu) \frac{r_c}{r} \sin \theta
-
2\nu \frac{r}{r_c} \sin \theta
+
(\frac{r_c}{r} - \frac{r_c^3}{r^3}) \sin 3\theta
\right),
\end{align}
where $(r, \theta)$ is the polar coordinate coresponding to the Cartesian coordinate $(x_1, x_2)$. In this problem, we use material properties $E=1 \text{GPa}$ and $\nu=0.3$.

In this problem, we pre-train the neural network based on the displacement and strain fields obtained from a coarse FEM solution with the mesh depicted in Fig. \ref{fig::2d-solid-hole-domain}(a). The loss function for the guided version of PINN is as follows:
\begin{equation}
\mathcal{L}^{\text{PINN+FEM}}_{\text{tot}}(\vec{\theta}) = 
\mathcal{L}^{\text{PINN}}_{\text{tot}}(\vec{\theta})
+
\mathcal{L}^{\text{FEM}}_{\text{tot}}(\vec{\theta}),
\end{equation}
where
\begin{equation}
\begin{split}
\mathcal{L}^{\text{FEM}}_{\text{tot}}(\vec{\theta})
&=
\mathcal{L}^{\text{FEM}}_{u}(\vec{\theta})
+
\mathcal{L}^{\text{FEM}}_{\epsilon}(\vec{\theta})
\\
&=
\frac{1}{N_{\text{FEM}}} \sum_{i=1}^{N_{\text{FEM}}} \norm{ \vec{u}^{\text{NN}}(\vec{x}_i^{\text{FEM}}; \vec{\theta}) - \vec{u}^{\text{FEM}}(\vec{x}_i^{\text{FEM}})}_2^2 
+
\frac{1}{N_{\text{FEM}}} \sum_{i=1}^{N_{\text{FEM}}} \norm{ \tensor{\epsilon}^{\text{NN}}(\vec{x}_i^{\text{FEM}}: \vec{\theta})  - \tensor{\epsilon}^{\text{FEM}}(\vec{x}_i^{\text{FEM}})}_{F}^2
\end{split},
\end{equation}
where $\vec{x}^{\text{FEM}}$ is the Cartesian coordinate of the observed FEM solution (Gauss quadrature points). The loss function corresponding to the FEM pre-training (i.e., $\mathcal{L}^{\text{FEM}}_{\text{tot}}(\vec{\theta})$) is discarded from the total loss function at training iteration 5000. The full batch training is performed by the \textit{Adam} optimizer with a learning rate 0.001. 
Also, \textit{ReduceLROnPlateau} learning scheduler of PyTorch is used with $\textit{patience}=50$, $\textit{factor}=0.8$, and $\textit{min\_lr}=1\mathrm{e}{-5}$. The same hyper-parameters are set for all the training tasks in this problem.

\begin{figure}[h]
 \centering
 \subfigure[Exact]
{\includegraphics[width=0.3\textwidth]{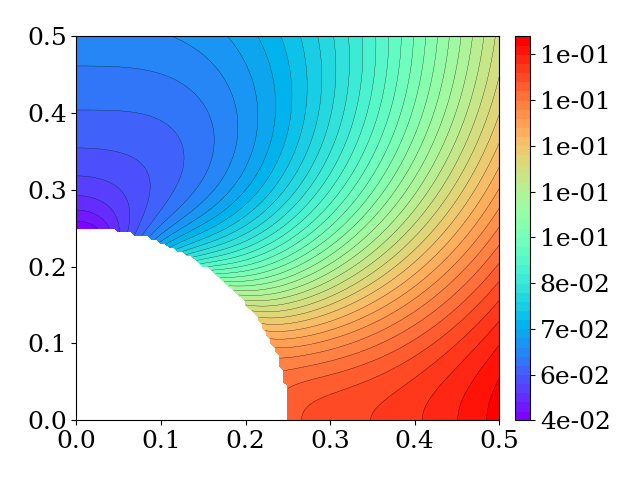}}
\hspace{0.01\textwidth}
 \subfigure[FEM+PINN+GradSurg]
{\includegraphics[width=0.3\textwidth]{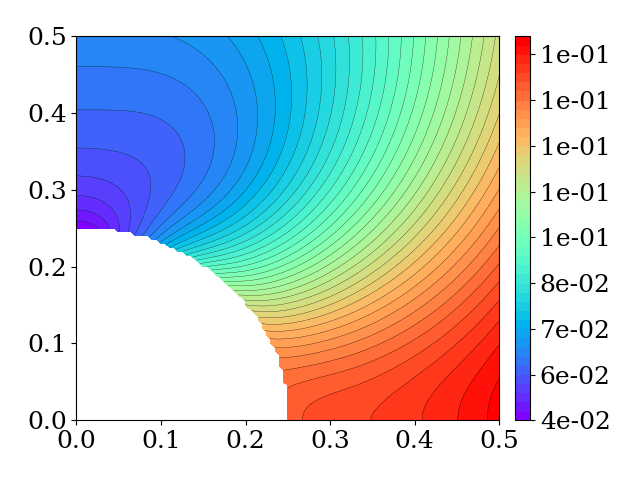}}
\hspace{0.01\textwidth}
 \subfigure[PINN]
{\includegraphics[width=0.3\textwidth]{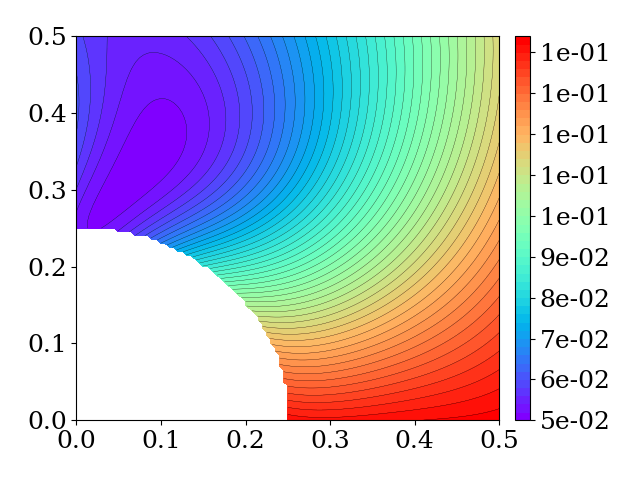}}
  \caption{ (a) contour of analytical solution of the displacement vector norm. (b) PINN solution guided by a coarse FEM solution; the optimizer is enhanced by the Gradient Surgery method to avoid gradient conflicts among different objectives defined in the total loss function. (c) plain PINN solution without any modification.
   \label{fig::2d-hole-disp-norm}}
\end{figure}

\begin{figure}[h]
 \centering
 \subfigure[Exact]
{\includegraphics[width=0.3\textwidth]{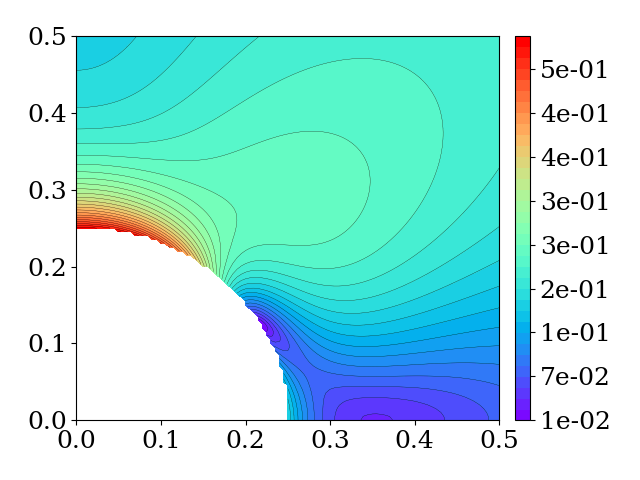}}
\hspace{0.01\textwidth}
 \subfigure[FEM+PINN+GradSurg]
{\includegraphics[width=0.3\textwidth]{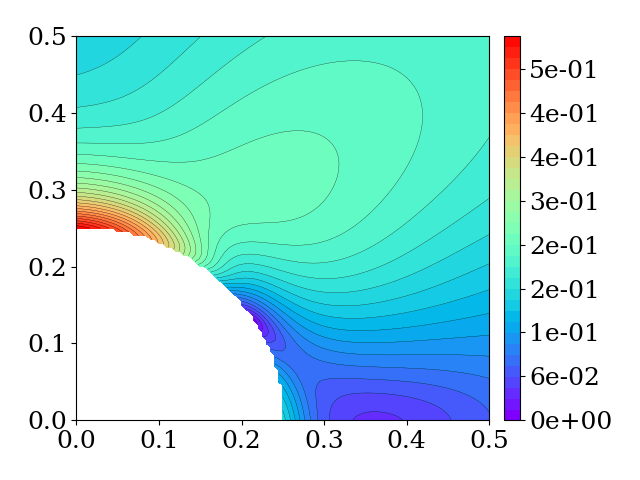}}
\hspace{0.01\textwidth}
 \subfigure[PINN]
{\includegraphics[width=0.3\textwidth]{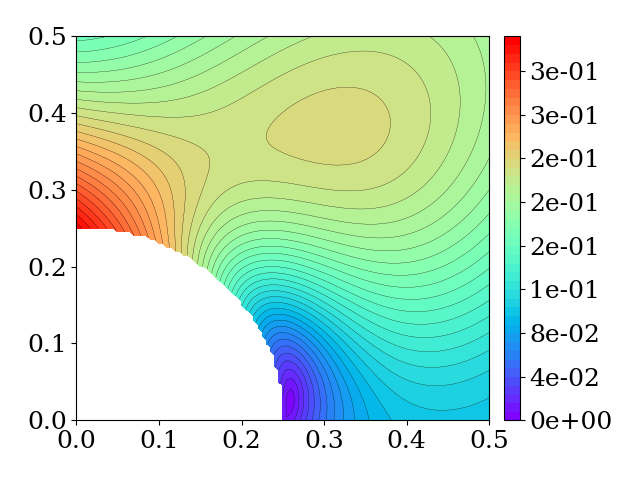}}
  \caption{ (a) analytical solution for the von Mises stress contour. (b) PINN solution guided by a coarse FEM solution; the optimizer is enhanced by the Gradient Surgery method to avoid gradient conflicts among different objectives defined in the total loss function. (c) plain PINN solution.
   \label{fig::2d-hole-von-mises}}
\end{figure}

\begin{figure}[h]
 \centering
 \subfigure[FEM+PINN+GradSurg]
{\includegraphics[width=0.35\textwidth]{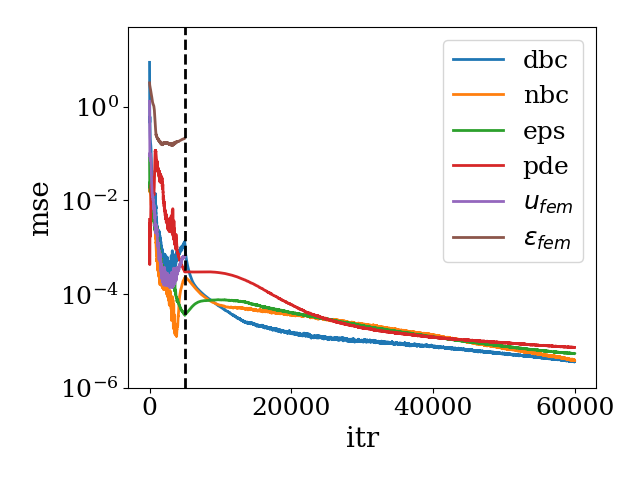}}
\hspace{0.01\textwidth}
 \subfigure[plain PINN]
{\includegraphics[width=0.35\textwidth]{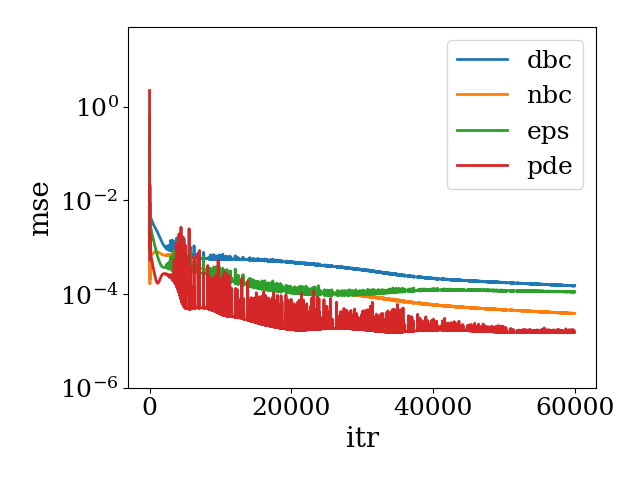}}
  \caption{ Mean square errors of each loss term during the iterations for (a) the enhanced scheme by FEM guidance and gradient surgery (b) plain PINN without any guidance and optimization treatments. The vertical dashed line in (a) indicates the iteration (i.e., iteration 5000) where the guidance from FEM is removed. The legends \emph{dbc}, \emph{nbc}, \emph{pde}, \emph{eps}, \emph{$u_{fem}$}, and \emph{$\epsilon_{fem}$} correspond to Dirichlet boundary condition, Neumann boundary condition, partial differential equation, strain compatibility, FEM displacement, and FEM strain loss terms, respectively.
   \label{fig::2d-hole-loss}}
\end{figure}

The results obtained by the proposed method and the plain PINN are compared in Figs. \ref{fig::2d-hole-disp-norm} and  \ref{fig::2d-hole-von-mises}. As the results suggest, the accuracy of the proposed method is better than the plain PINN approach. Notice that the plain PINN with appropriate weighted loss terms may improve results, however, these weights are not known apriori and are usually found by trial-and-errors as extra hyperparameters; such a greedy search over $N_{\text{obj}}=6$ real-valued weights exponentially increases the usage of computational resources. Nonetheless, our proposed approach based on the Gradient Surgery method does not introduce any extra hyperparameters. The mean square errors of each term in the loss function are reported in Fig. \ref{fig::2d-hole-loss}.

\subsection{Inverse two-dimensional anisotropic Poisson's problem}
Here, we showcase the application of proposed algorithms in dealing with the inverse problem where material properties are not known \textit{a priori}.
In this problem, the exact values of temperature and heat flux are given, and the goal is to estimate both of the conductivity tensor and PDE solution. The target conductivity tensor and analytical solutions are provided below:
\begin{align}
\tensor{K}_{\text{exact}} &= \begin{bmatrix}
1 & 1\\
1 & 2
\end{bmatrix},\\
 T_{\text{exact}}(\vec{x}) &= \sin (\pi x_1) \cdot \sin (\pi x_2).
\end{align}

The corresponding loss function, in this problem, has the following terms:
\begin{equation}
    \mathcal{L}_{\text{tot}} = 
    \mathcal{L}_{\text{pde}} + \mathcal{L}_{\text{dbc}} + \mathcal{L}_{T} +
    \mathcal{L}_{q_{1}} + \mathcal{L}_{q_{2}} +
    \mathcal{L}_{\text{PosDef}}
    \label{eq:loss-heat-inv}
\end{equation}
where $\mathcal{L}_{T}$ is the mse between neural network temperature prediction and provided exact solution, $\mathcal{L}_{q_{1}}$ and $\mathcal{L}_{q_{2}}$ are the mse between neural network flux prediction in the $x_1$ and $x_2$ directions, respectively.  To avoid nonphysical solutions for conductivity tensor, the positive definiteness of permeability tensor is also included in the loss function as an inductive bias towards the solution. Notice that the symmetric constraint of the conductivity tensor is explicitly induced by considering only the upper triangle components (i.e., $k_{11}$, $k_{12}$, and $k_{22}$) as the only degree of freedoms (extra trainable parameters). The loss term for the positive definiteness is chosen as follows:
%
%
\begin{equation}
 \mathcal{L}_{\text{PosDef}} =  \langle -\det(\tensor{K^{\text{approx}}}) \rangle , 
\end{equation}
where $\langle \cdot \rangle$ is the MacCauley bracket in which $\langle a \rangle=a$ if $a \geq 0$ and $\langle a \rangle=0$ otherwise, providing that $a$ is a real number. As such, $ \mathcal{L}_{\text{PosDef}}$ is activated  
only when the effective permeability obtained from the inverse problem is not positive semi-definite. 
The collocation points used in the \emph{pde} and \emph{PosDef} loss terms samples a structured grid of $30 \times 30$ points. Temperature and heat flux values on a structured grid of $45\times 45$ points are used in the loss terms $\mathcal{L}_T$, $\mathcal{L}_{q_1}$, and $\mathcal{L}_{q_2}$. 100 equidistant points are used for each boundary side.

We first train 20 teacher neural networks with the same hyperparameters where each network is initialized by the Xavier method with a different random seed. Teacher networks have 3 hidden layers with 10 units in each, and the hyperbolic tangent activation function is used.  Components of the conductivity tensor are randomly initialized from a uniform distribution $\tensor{K}^{\text{approx}} \sim \text{Uniform}(0, 1)$. These 20 teacher networks are trained by Adam optimizer and gradient surgery algorithm. Full batch training is performed by the learning rate 0.001 for 5000 iterations. Parameters $\textit{patience}=50$, $\textit{factor}=0.9$, and $\textit{min\_lr}=1\mathrm{e}{-6}$ are set for \textit{ReduceLROnPlateau} learning scheduler. After training these 20 teacher neural networks with the same hyperparameters, the best network is selected based on the metric Eq. \ref{eq:loss-heat-inv}. Then, the knowledge of the best-performed teacher is transferred to the student network with 100 hidden units in each hidden layer. The student network is trained with a learning rate equal to half of the teacher's learning rate.

\begin{figure}[h]
 \centering
 \subfigure[best teacher with 272 parameters]
{\includegraphics[width=0.3\textwidth]{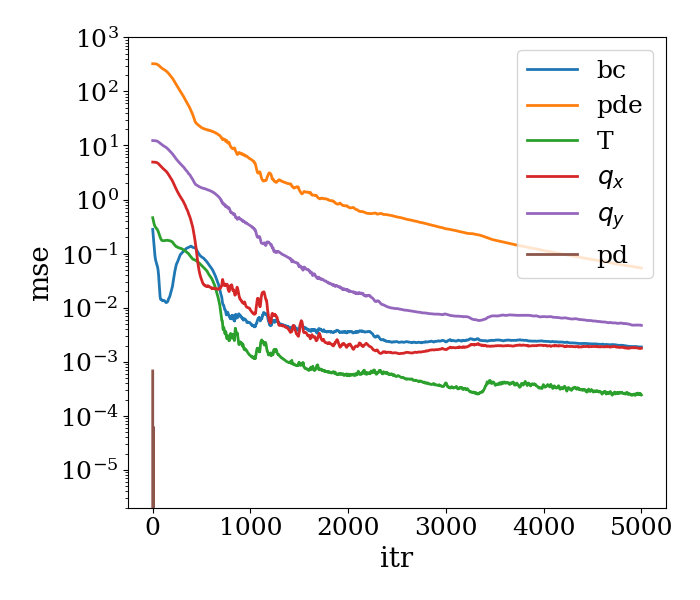}}
\hspace{0.01\textwidth}
 \subfigure[student with 20702 parameters]
{\includegraphics[width=0.3\textwidth]{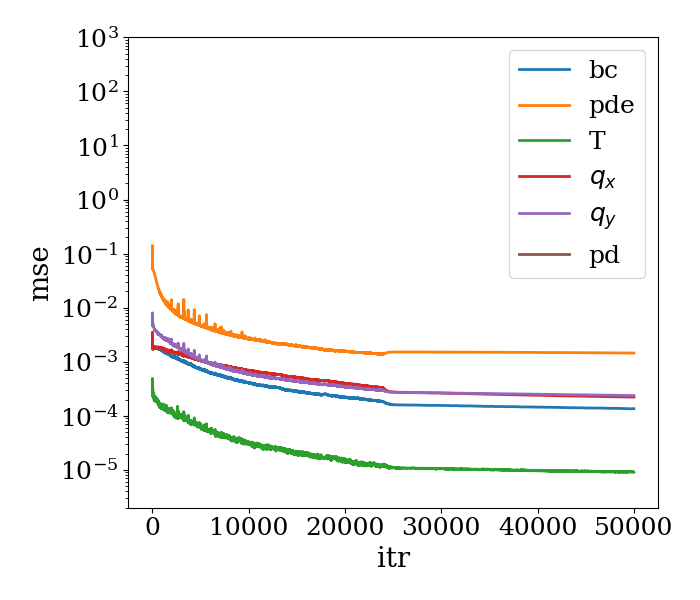}}
\hspace{0.01\textwidth}
 \subfigure[student with 20702 parameters]
{\includegraphics[width=0.3\textwidth]{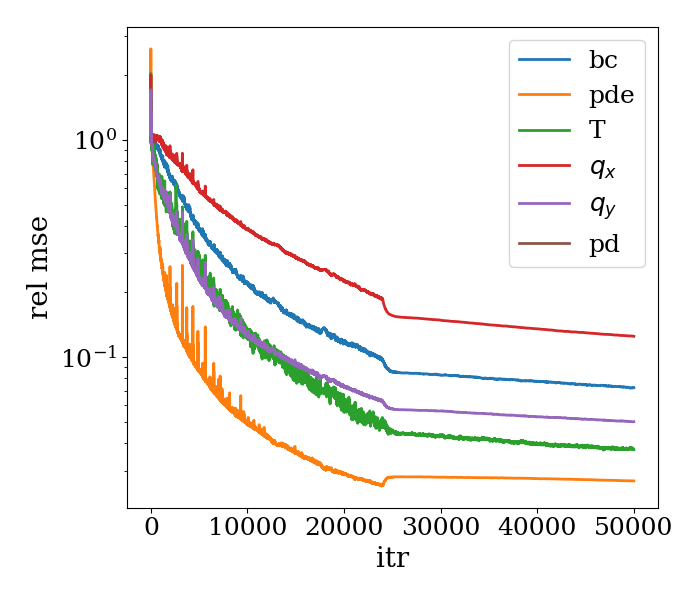}}
  \caption{ Mse of each loss term during the training for (a) the best-performed teacher network among 20 candidates, (b) student network.
  (c) relative mse of each loss term of the student network relative to the best-performed teacher network. 
  As the results suggest, the inherited knowledge from the teacher can preserve the level of each loss term at initial iterations (with only minor variations), and the loss terms are improved during the student network training. Legends \emph{bc}, \emph{pde}, \emph{T}, \emph{$q_x$}, \emph{$q_y$}, and \emph{pd} denote the corresponding loss terms to Dirichlet boundary condition, partial differential equation, temperature prediction, heat flux prediction in the x-direction, heat flux prediction in the y-direction, and positive definiteness of conductivity tensor. The loss term of \emph{pd} shown by the brown line is no-zero only for the few first iterations in Fig. (a); this is just observed for the best teacher network while for other networks this loss term may exist during all the iterations.
   \label{fig::2d-inv-poisson-loss}}
\end{figure}

\begin{figure}[h]
 \centering
 \subfigure[best teacher]
{\includegraphics[width=0.4\textwidth]{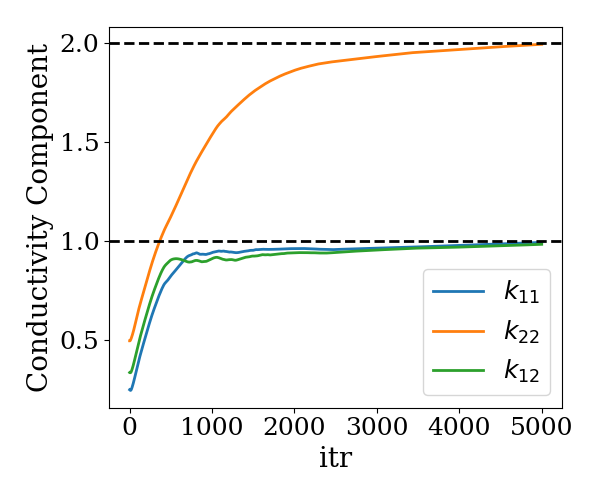}}
\hspace{0.01\textwidth}
 \subfigure[student]
{\includegraphics[width=0.4\textwidth]{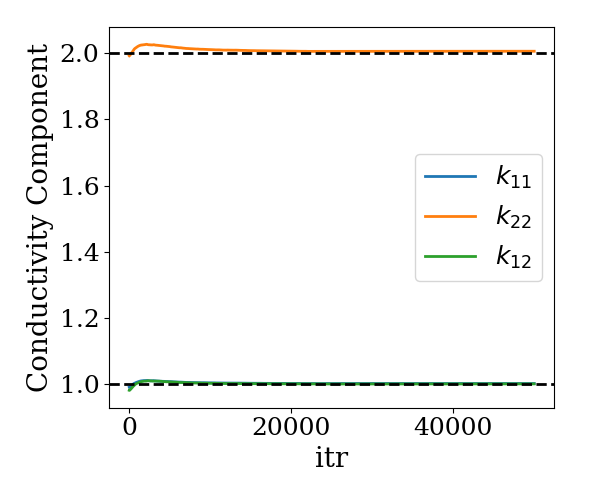}}
  \caption{ Conductivity tensor components during the iterations for (a) the best-performed teacher network, (b) the student network.
   \label{fig::2d-inv-poisson-perm}}
\end{figure}

\begin{figure}[h]
 \centering
 \subfigure[best teacher]
{\includegraphics[width=0.3\textwidth]{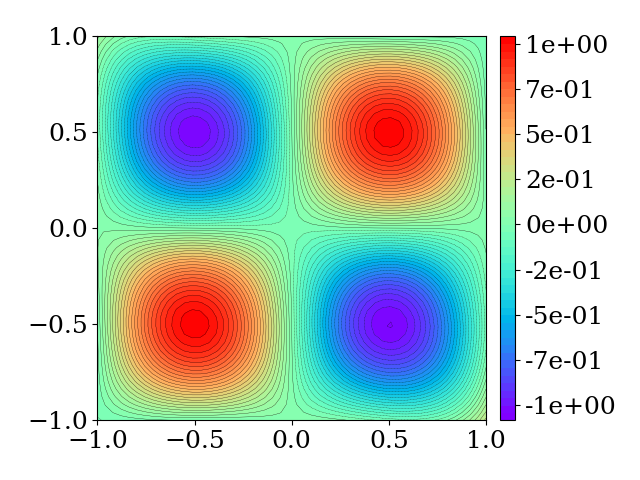}}
\hspace{0.01\textwidth}
 \subfigure[student]
{\includegraphics[width=0.3\textwidth]{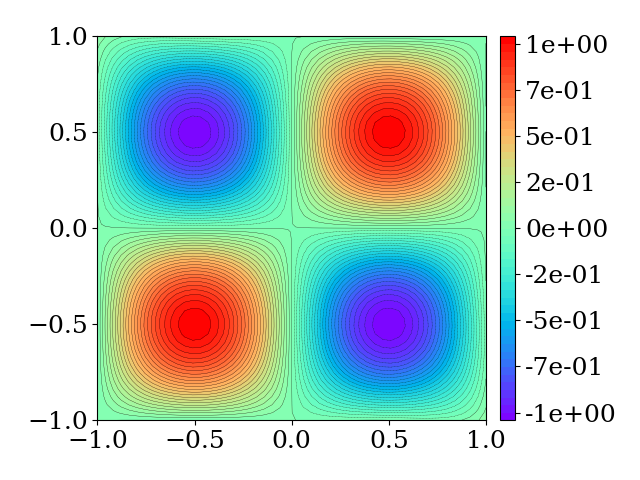}}
\hspace{0.01\textwidth}
 \subfigure[exact]
{\includegraphics[width=0.3\textwidth]{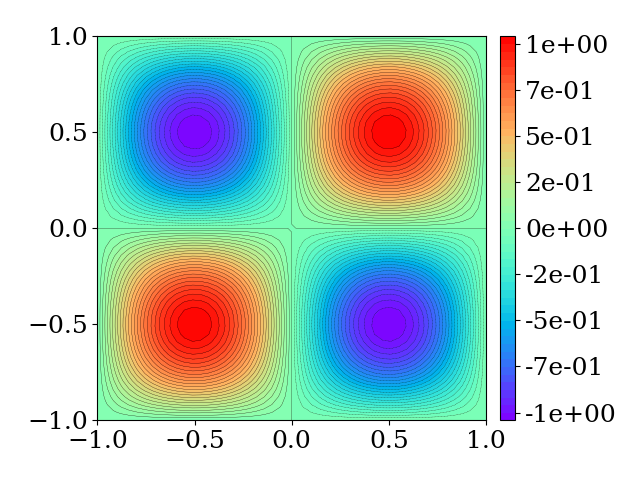}}
 \subfigure[difference between exact and best teacher]
{\includegraphics[width=0.3\textwidth]{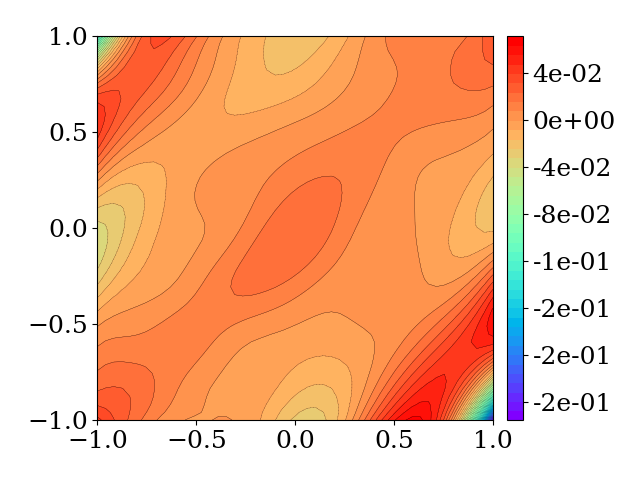}}
\hspace{0.1\textwidth}
 \subfigure[difference between exact and teacher]
{\includegraphics[width=0.3\textwidth]{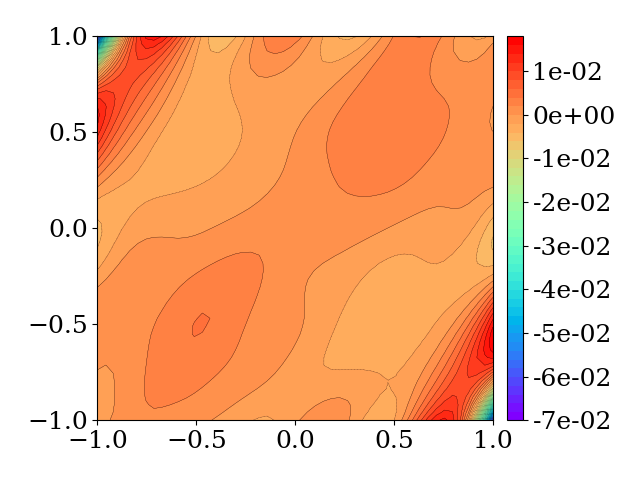}}
  \caption{ Temperature solution after training for (a) best-performed teacher network and (b) student network. 
  (c) exact solution, (d) difference between the exact solution and best-performed teacher network, and (e) difference between the exact solution and student network.
   \label{fig::2d-inv-poisson-sol-u}}
\end{figure}

\begin{figure}[h]
 \centering
 \subfigure[best teacher]
{\includegraphics[width=0.3\textwidth]{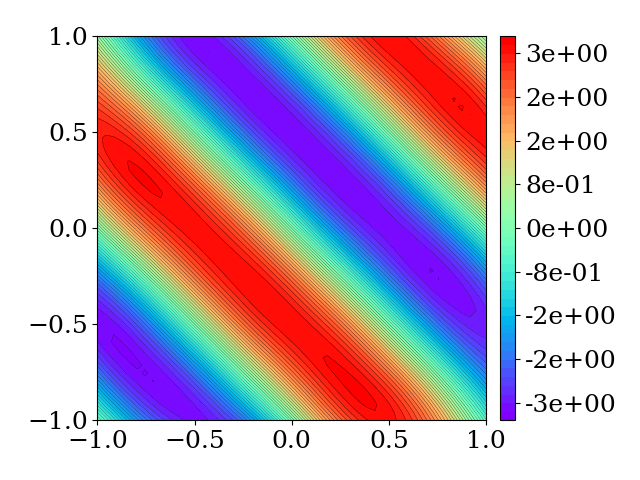}}
\hspace{0.01\textwidth}
 \subfigure[student]
{\includegraphics[width=0.3\textwidth]{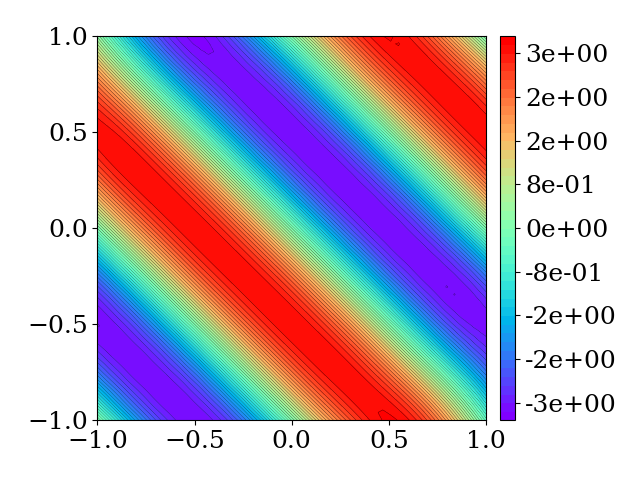}}
\hspace{0.01\textwidth}
 \subfigure[exact]
{\includegraphics[width=0.3\textwidth]{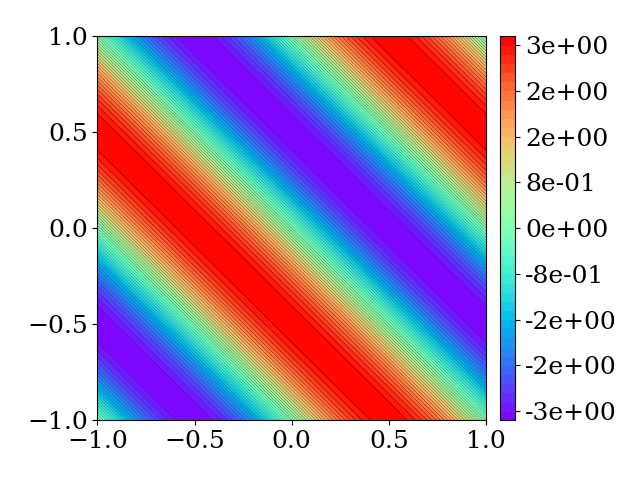}}
 \subfigure[difference between exact and teacher]
{\includegraphics[width=0.3\textwidth]{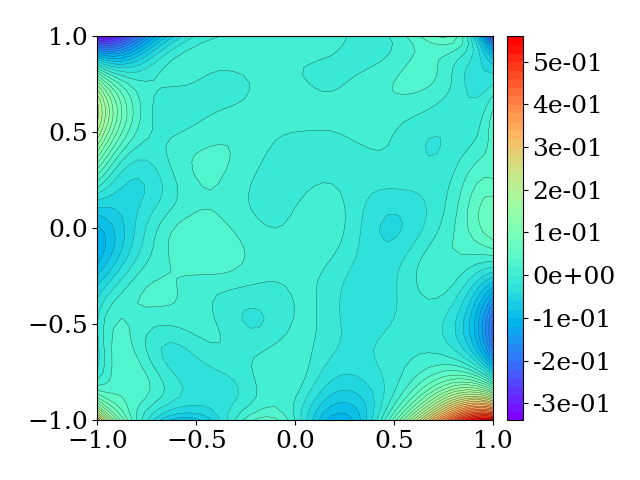}}
 \subfigure[difference between exact and student]
{\includegraphics[width=0.3\textwidth]{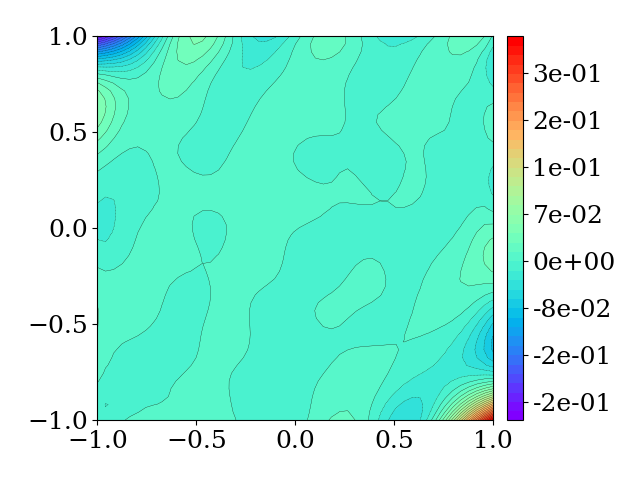}}
  \caption{Horizontal heat flux solution after training for (a) best-performed teacher network and (b) student network. 
  (c) exact solution, (d) difference between the exact solution and best-performed teacher network, and (e) difference between the exact solution and student network.
   \label{fig::2d-inv-poisson-sol-qx}}
\end{figure}

\begin{figure}[h]
 \centering
 \subfigure[best teacher]
{\includegraphics[width=0.3\textwidth]{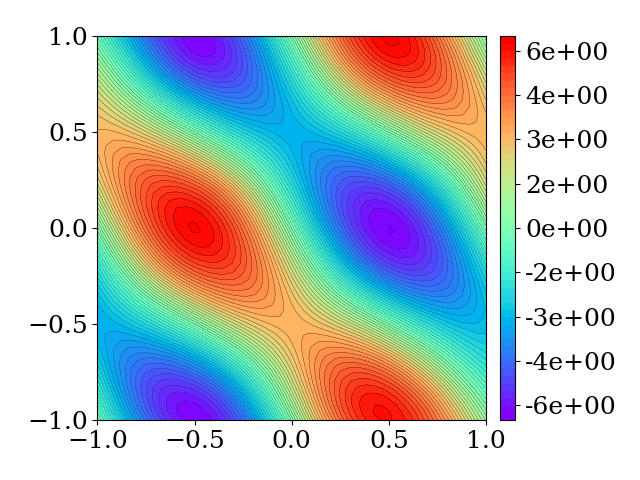}}
\hspace{0.01\textwidth}
 \subfigure[student]
{\includegraphics[width=0.3\textwidth]{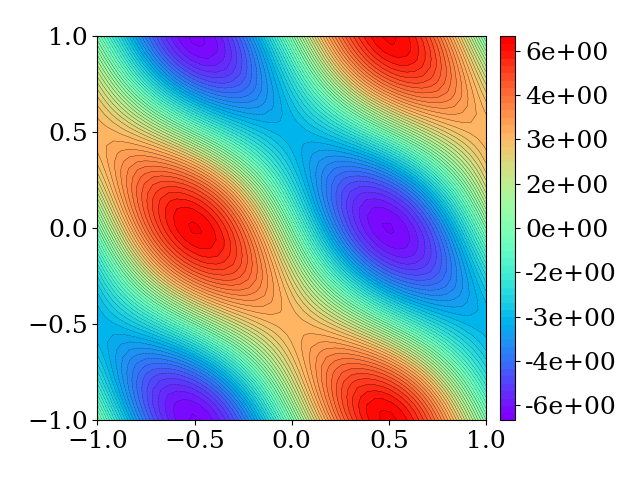}}
\hspace{0.01\textwidth}
 \subfigure[exact]
{\includegraphics[width=0.3\textwidth]{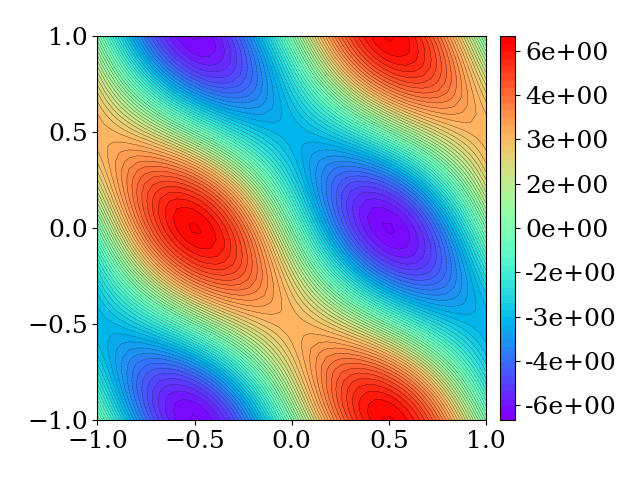}}
 \subfigure[exact - teacher]
{\includegraphics[width=0.3\textwidth]{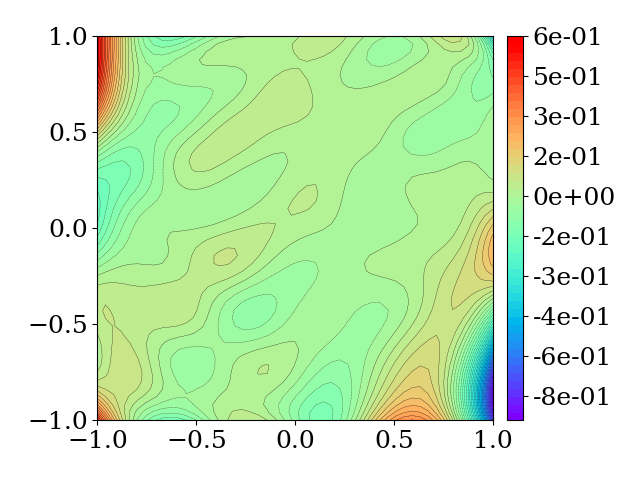}}
 \subfigure[exact - student]
{\includegraphics[width=0.3\textwidth]{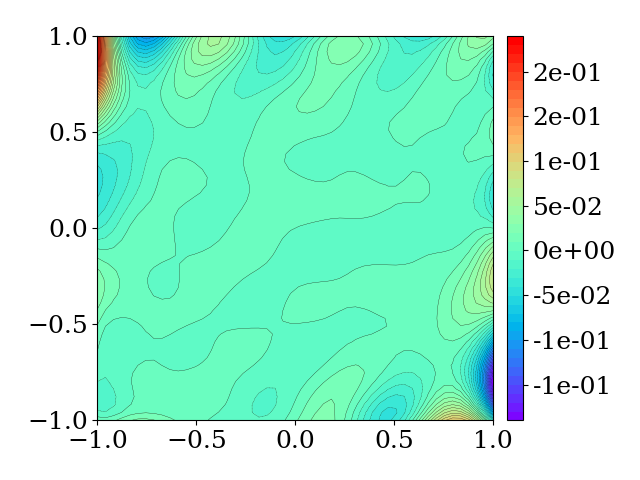}}
  \caption{Vertical heat flux solution after training for (a) best-performed teacher network and (b) student network. 
  (c) exact solution, (d) difference between the exact solution and best-performed teacher network, and (e) difference between the exact solution and student network.
   \label{fig::2d-inv-poisson-sol-qy}}
\end{figure}

Training performance of the best teacher and student is shown in Fig. \ref{fig::2d-inv-poisson-loss}. As the results indicate, although the number of parameters in the student neural network is increased by two orders of magnitude, the PINN training is stable and improving almost without loss of accuracy at the initial iterations. The recovered conductivity tensor by the teacher and student networks during the iterations are presented in Fig. \ref{fig::2d-inv-poisson-perm}. Again, as the results suggest, the near-optimal solution of the teacher network is almost close to a near-optimal solution of the student network, and the transferred knowledge remains stable during the student training. For a better comparison, the temperature and heat flux solutions are reported in Figs. \ref{fig::2d-inv-poisson-sol-u}, \ref{fig::2d-inv-poisson-sol-qx}, and \ref{fig::2d-inv-poisson-sol-qy}.

%
%
%
%

\section{Conclusions}
One of the major roadblocks on the adaptation of PINN for engineering problems is the hidden cost required to fine-tune the hyperparameters
to guarantee the learning rates and locate the global minima. For simpler task, finding the optimal way to configure 
the learning problems can be  handled by manual trial-and-error. 
However, for the PINN applications that exists a multitude of objectives, a manual trial-and-error based on intuition is insufficient. 
In this work, we attempt to provide a practical synthesis of algorithms and strategies that, while used together properly, 
may improve the robustness and performance of the PINN training. While the usage of spatial data to introduce supervised learning 
pre-training step may help us avoid to converge to a local minimum, the Net2Net can help us greatly simplify the weight initialization 
and ensure that the training of a high-resolution solution parametrized by a sizable neural network is feasible. 
Our results from benchmark experiments indicate that this
pre-training strategy, combined with the gradient survey that helps us addressing the conflicting gradients for different sub-tasks, 
shows great promises in easing the difficulty of training collocation PINN.

\section{Acknowledgments}
The authors are primarily supported by the  NSF CAREER grant from Mechanics of Materials and Structures program
 at National Science Foundation under grant contract CMMI-1846875 with additional support from 
 the Office of Advanced Cyber-infrastructure under grant contract OAC-1940203 and Air Force Office of Scientific Research 
 under grant contract FA9550-19-1-0318.
These supports are gratefully acknowledged. 
The views and conclusions contained in this document are those of the authors, 
and should not be interpreted as representing the official policies, either expressed or implied, 
of the sponsors, including the U.S. Government. 
The U.S. Government is authorized to reproduce and distribute reprints for 
Government purposes notwithstanding any copyright notation herein.

\begin{appendices}

\section{Boundary value problems used for numerical experiments} \label{sec:problemstatement}
In this work, we apply our proposed training procedure to forward and inverse problems governed by elliptic PDEs, although the methods are not restricted to this type of PDEs. For completeness, we briefly review the two set of boundary value problems 
the heat conduction (Poisson's equation) and static elasticity problems in 1D and 2D we used in the numerical experiments showcased in Section \ref{sec:numric}.

\subsection{Heat conduction}
The steady-state heat conduction in $d=2,3$ dimensional space can be described by \citet{kaviany2012principles}:
\begin{align}
&\nabla_{x} \cdot \vec{q(\vec{x})} + s(\vec{x}) = 0; \ \vec{x} \in \Omega,\label{eq:poission-pde}\\
&\vec{q}(\vec{x}) = \tensor{K}(\vec{x}) \cdot \nabla_{x} T(\vec{x}); \ \vec{x} \in \Omega,\label{eq:poission-const-law}\\
&T(\vec{x}) = \bar{T}(\vec{x}); \ \vec{x} \in \partial \Omega_T,\\
&\vec{q}(\vec{x}) \cdot \vec{n}(\vec{x}) = \bar{q}(\vec{x}); \ \vec{x} \in \partial \Omega_q,
\label{eq:poission}
\end{align}
where $T$ is the temperature field, $\vec{q}$ is the heat flux vector, $\tensor{K}$ is the conductivity tensor, $s$ is the source/sink term, $\vec{x}$ is the spatial coordinate, $\Omega \subset \mathbb{R}^d$ is the domain, $\partial \Omega_T \subset \mathbb{R}^{d-1}$ is the Dirichlet boundary surface, $\partial \Omega_q \subset \mathbb{R}^{d-1}$ is the Neumann boundary surface, $\vec{n}$ is the outward normal vector over domain surface $\partial \Omega \subset \mathbb{R}^{d-1}$. The above PDE is well-posed if the conductivity tensor is positive-definite and $\partial \Omega_T \cap \partial \Omega_q = \emptyset, \ \partial \Omega_T \cup \partial \Omega_q = \partial \Omega$.

\subsection{Elasticity}
Neglecting the inertia effect, the deformation of an elastic body is described by \cite{malvern1969introduction}:
\begin{align}
&\nabla_{x} \cdot \tensor{\sigma(\vec{x})} + \vec{b}(\vec{x}) = \vec{0}; \ \vec{x} \in \Omega,\label{eq:elasticity-pde}\\
&\tensor{\sigma}(\vec{x}) = \mathbb{C}(\vec{x}) : \tensor{\epsilon}(\vec{x}); \ \vec{x} \in \Omega,\label{eq:elasticity-const-law}\\
&\tensor{\epsilon}(\vec{x}) = \nabla_{x}^{\text{sym}} \vec{u}(\vec{x}); \ \vec{x} \in \Omega,\label{eq:elasticity-compability}\\
&\vec{u}(\vec{x}) = \bar{\vec{u}}(\vec{x}); \ \vec{x} \in \partial \Omega_u,\\
&\tensor{\sigma}(\vec{x}) \cdot \vec{n}(\vec{x}) = \bar{\vec{t}}(\vec{x}); \ \vec{x} \in \partial \Omega_{\sigma},
\end{align}
where $\vec{u}$ is displacement vector, $\vec{b}$ is body force vector, $\tensor{\sigma}$ is stress rank two tensor, $\mathbb{C}$ is the stiffness rank four tensor with minor and major symmetries, $\tensor{\epsilon}$ is the strain rank two tensor, $\partial \Omega_u \subset \mathbb{R}^{d-1}$ is the Dirichlet boundary surface where prescribed displacement $\bar{\vec{u}}$ is known, $\partial \Omega_{\sigma} \subset \mathbb{R}^{d-1}$ is the Neumann boundary surface where prescribed traction $\bar{\vec{t}}$ is known. The above PDE is well-posed if the stiffness tensor is positive-definite and $\partial \Omega_u \cap \partial \Omega_{\sigma} = \emptyset, \ \partial \Omega_u \cup \partial \Omega_{\sigma} = \partial \Omega$.

\section{1D Poisson's problem for comparing adaptive guided PINN with plain PINN}\label{appx:toy1}
In the first 1D Poisson problem, there is no contribution from the Neumann boundary condition, and other loss terms in the plain PINN setup has the following form:
\begin{align}
    \mathcal{L}_{\tpde}(\vec{\theta}) &= \frac{1}{N_{\tpde}} \sum_{i=1}^{N_{\tpde}} 
                          \left(
                            \frac{\partial q^{\tnn}(x_i^{\tpde}; \vec{\theta})}{\partial x} + s(x_i^{\tpde})
                          \right)^2\label{eq:toy-1-pde},\\
    \mathcal{L}_{\tdbc}(\vec{\theta}) &= \frac{1}{N_{\tdbc}} \sum_{i=1}^{N_{\tdbc}} 
                          \left(
                             T^{\tnn}(x_i^{\tdbc}; \vec{\theta}) - \bar{T}(x_i^{\tdbc})
                          \right)^2 \label{eq:toy-1-dbc}.
\end{align}
The neural network approximates only the temperature field $T^{\tnn}$, and the heat flux is calculated via the automatic differentiation $q^{\tnn} = \partial T^{\tnn} / \partial x$. 60 equidistant collocation points $\{ x_i^{\tpde} \}_{i=1}^{60} \in (-10, 10)$ are used for the PDE contribution and $x_1^{\tdbc} = -10, x_2^{\tdbc} = 10$.

In this problem, there exists only one type of auxiliary information which is a coarse FEM temperature solution at nodal points:
\begin{equation}
    \mathcal{L}_1^{\taux}(\vec{\theta}) = \mathcal{L}_{T}^{\taux}(\vec{\theta}) = 
    \frac{1}{N_{\taux1}} \sum_{i=1}^{N_{\taux1}} 
                          \left(
                            T^{\tnn}(x_i^{\taux1}; \vec{\theta}) - T^{\tfem}(x_i^{\taux1})
                          \right)^2
\end{equation}

\section{1D Poisson's problem for the illustration of Net2Net transfer learning in PINN}\label{appx:toy2}
In this problem, we aim to find the solution of a 1D heat conduction with the following manufactured solution by Net2Net transfer learning:
\begin{equation}
    T_{\text{exact}}(x) = \sin(1.2\pi x); \quad x \in [-1, 1].
\end{equation}
The problem is formulated in the mixed form where the neural network output includes the temperature and heat flux, i.e., $\mathcal{F}_{\vec{\theta}}(x) = [T^\tnn, q^\tnn]^T$. The PDE and DBC contributions are the same as Eqs. \ref{eq:toy-1-pde} and \ref{eq:toy-1-dbc}. The loss term corresponding to the compatibility (or constitutive law) constraint is defined as:
\begin{equation}
    \mathcal{L}_{\tcompy}(\vec{\theta}) = \frac{1}{N_{\tcompy}} \sum_{i=1}^{N_{\tcompy}} 
                          \left(
                            \frac{\partial T^{\tnn}(x_i^{\tcompy}; \vec{\theta})}{\partial x} - q^{\tnn}(x_i^{\tcompy})
                          \right)^2.
\end{equation}
In this problem, the set of collocation points for PDE and compatibility constraints is chosen the same and discretizes the domain with equidistant points as $\{ x_i^{\tpde} \}_{i=1}^{100} = \{ x_i^{\tcompy} \}_{i=1}^{100} \in (-1, 1)$. This problem focuses only on the Net2Net idea, hence there is no auxiliary task, also the gradient surgery is not utilized.

\end{appendices}

\bibliographystyle{plainnat}
\bibliography{main}

\end{document}